\documentclass{article}
\usepackage{arxiv}

\usepackage[utf8]{inputenc} 
\usepackage[T1]{fontenc}    
\usepackage{hyperref}       
\usepackage{url}            
\usepackage{booktabs}       
\usepackage{amsfonts}       
\usepackage{nicefrac}       
\usepackage{microtype}      
\usepackage{lipsum}

\usepackage{graphicx}
\usepackage{amssymb}
\usepackage{amsthm}
\usepackage{amsmath}

\usepackage[numbers,sort&compress ]{natbib}
\usepackage{float}
\usepackage[table]{xcolor}
\usepackage{xcolor}
\usepackage{pdflscape}
\usepackage{multirow}
\usepackage{subcaption}
\usepackage{array}
\newcolumntype{C}[1]{>{\centering\arraybackslash}p{#1}}

\title{Local region-learning modules for point cloud classification}

\author{
  Kaya Turgut \\
  Department of Electrical-Electronics Engineering\\
  Eskisehir Osmangazi University\\
  Eskisehir, Turkey \\
  \texttt{kayaturgut@hotmail.com} \\
   \And
  Helin Dutagaci \\
  Department of Electrical-Electronics Engineering\\
  Eskisehir Osmangazi University\\
  Eskisehir, Turkey \\
  \texttt{hdutagaci@ogu.edu.tr} }

\begin{document}
\maketitle

\begin{abstract}
Data organization via forming local regions is an integral part of deep learning networks that process 3D point clouds in a hierarchical manner. At each level, the point cloud is sampled to extract representative points and these points are used to be centers of local regions. The organization of local regions is of considerable importance since it determines the location and size of the receptive field at a particular layer of feature aggregation. In this paper, we present two local region-learning modules: Center Shift Module to infer the appropriate shift for each center point, and Radius Update Module to alter the radius of each local region. The parameters of the modules are learned through optimizing the loss associated with the particular task within an end-to-end network. We present alternatives for these modules through various ways of modeling the interactions of the features and locations of 3D points in the point cloud. We integrated both modules independently and together to the PointNet++ and PointCNN object classification architectures, and demonstrated that the modules contributed to a significant increase in classification accuracy for the ScanObjectNN data set consisting of scans of real-world objects. Our further experiments on ShapeNet data set showed that the modules are also effective on 3D CAD models.
\end{abstract}


\section{Introduction}
\label{section:intro}

Analysis and semantic interpretation of 3D point clouds became a popular research topic due to the widespread use of 3D sensors for many public, commercial and scientific activities. Many 3D sensors provide output in the form of point clouds. In order to process these 3D point clouds for various tasks, point-based deep neural network architectures have been developed \citep{Bello2020Survey, Guo2021Survey, Liu2019Survey}, following the unrivaled success of deep learning methods in pattern recognition and computer vision.

One of the main challenges such 3D point-based deep learning approaches face is the lack of inherent indexing structure that defines the spatial relationships between data points. A point cloud is an unordered set of points, each independently represented by their Cartesian coordinates. As opposed to the grid-based indexing, which provides information on spatial relationships between pixels in images, the spatial relationships between points in the point cloud have to be explicitly specified.

In order to define spatial relationships within the point cloud, some deep learning architectures involve organization of the point data in a hierarchical manner \citep{Qi2017PointNet2, Li2018PointCNN, Thomas2019KPConv}. At each organizational layer, the point cloud is subsampled to obtain representative points (sampling), then a local region around each representative point is defined (grouping) for aggregation of point features within this region. Similar to the spirit of Convolutional Neural Networks (CNN), the number of representative points is reduced at each layer and their corresponding receptive fields are widened. While the features of numerous points in the first layers encode local surface information, the features of fewer points in subsequent layers represent more of the global structure.  

The organization of the point cloud in such hierarchical manner depends on two design choices: 1) Determination of the representative points, and 2) The size of the neighborhood defined by the representative points at each layer. In PointNet++ \citep{Qi2017PointNet2}, which is one of the first 3D point-based deep learning approaches, the representative points are selected using Farthest Point Sampling (FPS), and the neighborhood size is fixed to a layer-specific radius value for all representative points. In PointCNN \citep{Li2018PointCNN}, another architecture that organizes point clouds hierarchically, representative points are selected using random sampling. Dilated $K$ nearest neighbor search is employed for grouping, where the dilation factor and $K$ control the size of receptive fields. In order for the network to rearrange the locations and sizes of receptive fields in a task-dependent manner, we propose two local region-learning modules: 1) Center Shift Module (CSM), and 2) Radius Update Module (RUM).   

Farthest point sampling, random sampling, and grid sampling are widely used techniques to define representative points that cover the point cloud. We propose the employment of the Center Shift Module to infer the appropriate replacement of each representative point using the local information surrounding it. Similarly, Radius Update Module, as opposed to using a layer-specific fixed radius for grouping around all center points, alters the radius (i.e. the size of the receptive field) for each center point. CSM and RUM learn to infer the replacement of the points and the radii of their receptive fields, respectively, through training with loss functions specified for a particular task; object classification in our study.

Our main contribution is the development of two novel modules that learn to determine the location and size of local regions through training to maximize the classification accuracy of point cloud objects. The two modules can be integrated into any hierarchical point-based deep learning network, where, at each layer, the points are sampled into representative points and grouped within spherical receptive fields around these points. For each module, we propose alternative approaches for inferring the effective locality of each point. These variations correspond to the manner of point interactions within the point cloud. We especially drew inspiration from attention-based approaches \citep{Vaswani2017AttYNeed, Zhao2020ExpSA, Guo2021PCT, Zhao2020PT} to encode these interactions.

\section{Related Work}
\label{section:related_work}

In this section, we discuss methods that form local regions on point clouds, alternative to common practices such as FPS, random sampling, ball query and KNN search, for deep learning architectures. These methods are mainly divided into two categories: 1) Methods that determine the representative points, i.e., centers of local regions, and 2) methods that group the local points around center points. We discuss these lines of work in the following subsections.

\subsection{Determining Centers of Local Regions}
The heuristic method FPS has been widely used to determine the center points of local regions. Since the points obtained by FPS are distributed uniformly in the Euclidean space, full coverage of the point cloud is provided. However, the selected points do not necessarily correspond to the distinctive parts of objects. To address this problem, two approaches have been proposed: Pre-processing of the point cloud and processing within the end-to-end deep learning framework dedicated to the particular task.

In the first approach, centers of local regions are determined independently from the main network that performs classification or segmentation of the point cloud. In \citep{Li2018SoNet}, a self-organized map (SOM) is trained to produce a two-dimensional representation of the input point cloud in terms of $m\times m$ nodes. Initially, a fixed number of nodes are selected randomly and the final locations of these nodes are determined via unsupervised competitive learning. Zhang and Jin \citep{Zhang2022AOMC} proposed an adaptive clustering method to prepare the original point cloud for the deep learning network. They determined the number of clustering centers according to the change of the within-cluster sum of squared errors (SSE). Sampling centers are obtained through forcing the mutual distance between initial cluster centers as far apart as possible. In \citep{Landrieu2018SP}, partitioning the point cloud into geometrically homogeneous regions in the form of a superpoint graph is proposed. In Grid-GCN \citep{Xu2020GridGCN}, a Coverage-Aware Grid Query (CAGQ) module is introduced. This module consists of a center voxels sampling framework, and a node points querying algorithm to pick $K$ points around each center point.  

The parameters of the data organization methods described above are not optimized jointly with the parameters of the task network. In the second type of approach, parameters of the procedure that determines the center points are learned through an end-to-end deep learning architecture. In \citep{Dovrat2019SNET}, a sampling network called S-NET is proposed. During training, S-NET is connected to a pre-trained and fixed task network, and the parameters of S-NET are learned through minimization of the task's loss and a sampling loss. However, in the inference stage, the sampled points need to be matched with the original point cloud. Lang et al. \cite{Lang2020SampleNet} extended S-Net \cite{Dovrat2019SNET} to integrate the matching step into the learning process by using soft projection. In \cite{Yang2019Gumbel}, Gumbel Subset Sampling (GSS) is proposed to select representative points within the end-to-end task-dependent network. In CP-Net \citep{Nezhadarya2020CPNet}, Critical Points Layer (CPL) is introduced to keep representative points, according to a point’s level of contribution to the
global max-pooling. PointASNL \cite{Yan2020PointASNL} is proposed to reduce the effect of noise-induced points. The coordinates and attributes of the center points, which are initially sampled by the FPS algorithm, are updated using a self-attention approach. In SK-Net \citep{Wu2020SKNetDL}, unlike SO-Net \cite{Li2018SoNet}, the nodes are jointly optimized with regression over extracted features by the end-to-end framework. In PointDAN \cite{Qin2019PointDAN}, designed for domain adaptation, center points are determined initially by FPS and the shift amounts of the center points are learned by a sub-network. Lin et al. \citep{Lin2021DANet} introduced a density-adaptive deep learning framework called DA-Net. Initial center points are learned in relation to the aggregated feature in a global manner and then, they are shifted with a density adaptive sampling module to ease the effect of noise.

\subsection{Grouping Local Points}

Determination of adaptive receptive fields based on geometric similarities within the point cloud is a challenging problem. Similar to the determination of sampling points, adaptive adjusting of the grouping process can be done independently or learned within the task network. As an example for the first approach, Sheshappanavar and Kambhamettu \citep{Sheshappanavar2021DLocal} proposed a statistical method based on the distribution of local points. The receptive field is considered as an ellipsoid and its size and orientation are calculated using the eigenvalues and eigenvectors of covariance analysis within the defined initial local region. 

There are also works that modify the receptive fields around the center points within the task network. Inspired by dilated convolution, Qiu et al. \citep{Qiu2021DenseResolutionNF} proposed the Adaptive Dilated Point Grouping module. The dilation factor for nearest neighborhood algorithm is learned by a network over local regions determined with maximum dilation factor. In another work \citep{Wang2020AdapRecpField}, instead of learning the dilation factor, local regions are constituted by a nearest neighbor search algorithm with a dilation factor. The adaptive feature of regions are extracted over local points through learned weight parameters. The attention search module proposed in  \citep{Xiang2022ATSearch} uses an adaptive combination of spatial and feature distances over nearest neighbours. The weights on spatial and feature distances are learned by an attention mechanism and thus the receptive field is altered implicitly for nearest search algorithm. 

In order to jointly define representative points and their local neighborhoods through learned parameters, the Spatial Location Feature Transform Function was proposed in \cite{Li2022PSNet}. The dimension of the resulting features of the transform is constrained to the desired number of center points. The channels of the features of the point cloud are sorted independently in descending order and the point index at the top of each sorted channel is selected as a center point. The following $K$ points at each sorted channel constituted the local region around the corresponding center point.

\begin{figure}[h]
     \centering
     \includegraphics[width=1.0\textwidth]{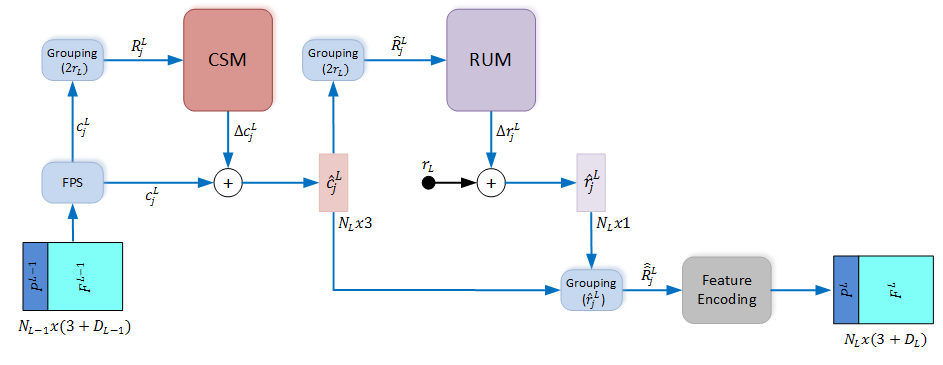} \caption{Integration of CSM and RUM modules into  a hierarchical point-based deep learning framework. }
     \label{fig:integration_CSM_RUM}
\end{figure}

In our work, we propose modules that provide adaptive updates for both the locations of center points and the sizes of the receptive fields around them. These modules can operate within any end-to-end network necessitating point sampling and grouping. The parameters that infer the local regions are learned through minimization of a combination of the task-dependent loss function and a loss function that limits the amount of alterations on center points and radii. We designed most variations of our Center Shift Module to be heavily influenced by attention-based modeling of point interactions. Also, the proposed Radius Update Module is different from the above mentioned approaches in that it directly increases or decreases the radius of the receptive field for each center point depending on point feature interactions and spatial organization of points in the locality.

\section{Method}
\label{section:method}

In 2D CNNs, at each layer, the size of the image is reduced by pooling while the effective region of each convolution is increased. This approach is mimicked for point clouds through sampling and grouping operations. In Figure \ref{fig:integration_CSM_RUM}, a representative layer for a hierarchically-organized 3D point-based deep network is given together with the alterations provided by CSM and RUM. At each layer $L$, $N_L$ new representative points are sampled from the $N_{L-1}$ representative points of the previous layer $L-1$, with $N_L < N_{L-1}$. Previous representative points are grouped around new representative points, which can also be referred to as "center points". 

Let the set of representative points corresponding to layer $L-1$ be denoted by $\mathcal{P}^{L-1}=\{p_1^{L-1},p_2^{L-1},...,p_{N_{L-1}}^{L-1} \}$, $p_i^{L-1}\in\mathbb{R}^3 $, and their corresponding point features be in the set $\mathcal{F}^{L-1}=\{f_1^{L-1},f_2^{L-1},...,f_{N_{L-1}}^{L-1}\}$, $f_i^{L-1}\in\mathbb{R}^{D_{L-1}}$. Through FPS, a subset of $\mathcal{P}^{L-1}$ is formed as $\mathcal{C}^{L}=\{c_1^L,c_2^L,...,c_{N_{L}}^L \}$, $ c_j^L\in\mathbb{R}^3 $, with corresponding features $\mathcal{G}^{L}=\{g_1^{L},g_2^{L},...,g_{N_{L}}^{L}\}$, $ g_j^{L}\in\mathbb{R}^{D_{L-1}}$, where $\mathcal{G}^{L} \subset \mathcal{F}^{L-1}$. Let $\mathcal{R}_j^{L}=\{p_{j,1},p_{j,2},...,p_{j,K}\}$, with $p_{j,k} \in \mathcal{P}^{L-1}$ denote the set of $K$ points randomly selected from the spherical region, centered at $c_j^L$, with a specific radius. Without the proposed Center Shift Module (CSM) and Radius Update Module (RUM), a hierarchical deep learning architecture aggregates the features of these $K$ points into the region specific feature $f_j^L\in\mathbb{R}^{D_L}$, attached to $c_j^L$, to be forwarded to the next layer. The procedure is repeated in the next layer with $\mathcal{P}^L$ = $\mathcal{C}^{L}$, and the corresponding feature set $\mathcal{F}^L$.

With the integration of CSM and RUM, $K$ points in $\mathcal{R}_j^{L}$ are incorporated into the task of updating the locations of sampled center points and the radii of the groupings through examining point interactions. After each update, $\mathcal{R}_j^{L}$ is also updated accordingly. As seen in Figure \ref{fig:integration_CSM_RUM}, once initial center points $c_{j}^{L}$ are obtained through FPS algorithm, the task of CSM is to infer the shift vector $\Delta c_{j}^{L}$ for each center point so that the center point is updated as:

\begin{equation} \label{eq:hatcj}
   \hat{c}_{j}^{L} = c_{j}^{L} + \Delta c_{j}^{L}
\end{equation}

RUM is responsible of determining $\Delta r_{j}^{L}$ for each center point. The receptive field of the $j$th center point is thus updated as:

\begin{equation} \label{eq:radius}
   \hat{r}_{j}^{L} = r^{L} + \Delta r_{j}^{L}
\end{equation}

We suggest a number of variants for the design of both CSM and RUM. The details of these variants are given in the following subsections. Unless necessary, the superscript $L$, indicating the layer is dropped from the notation for the sake of simplicity.

\subsection{Center shift module}
\label{sec:CSM}

The objective of CSM is to move the center points sampled by FPS algorithm using local and/or global organization and interactions of other points and their features. Through various ways of encoding these interactions, we implemented five variations of CSM.

\subsubsection{CSM-I}

CSM-I is similar to the geometrically guided shift learning proposed by \citet{Qin2019PointDAN} in that the edges between the point to be shifted and its neighbors are weighted by a transformation of the feature relations. Our contribution is to update the feature of the center point with an attention-based mechanism, and determine the edge weights through the difference between the features of the surrounding points and the aggregated feature vector of the center point.

Given an initial center point, $c_j$ with point feature vector $g_j$, let $\mathcal{R}_j$ be the local region centered at $c_j$. The radius of the spherical region $\mathcal{R}_j$ is set to be $2r^L$. $K$ points are randomly selected from this region defining the neighboring points $p_{j,k}$ with point feature vectors $f_{j,k}$. We define the displacement vector affected by point $p_{j,k}$ on $c_j$ as:

\begin{equation} \label{eq:deltaEjk}
   \Delta E_{j,k} = \gamma (\hat{g}_j - f_{j,k}) \odot (c_{j} - p_{j,k}), \;\; k = 1, ..., K
\end{equation}

\noindent where $\odot$ represents the Hadamard product, and $\hat{g}_j$ is the updated feature vector of the center point. The displacement vector $\Delta E_{j,k} \in \mathbb{R}^3$ is a weighted version of the relative location vector between the center point, $c_{j}$ and the neighbor point, $p_{j,k}$. The weights are determined by the 3D output of $\gamma$, which is an MLP network with two hidden layers, operating on the difference between the updated feature vector of the center point, $\hat{g}_j$ and the feature vector, $f_{j,k}$ of the neighbor point. ReLU is used as the activation function of the first hidden layer, while $tanh$ is preferred for the second hidden layer to limit the size of the displacement vector. The final displacement vector $\Delta c_{j}$ is obtained through averaging the displacement vectors affected by $K$ neighbor points:

\begin{equation} \label{eq:deltacj}
   \Delta c_{j}= \frac{1}{K} \sum_{k=1}^{K} \Delta E_{j,k}
\end{equation}

Notice that, instead of using the feature vector, $g_j$, transmitted from the previous layer, we use an updated version of it, $\hat{g}_j$, for assessing the center point's similarity to the feature vectors of the neighbor points:

\begin{equation} \label{eq:hatg}
   \hat{g}_j = g_j + g_j^{sa} 
\end{equation}

The vector $g_j^{sa}$ is an attention-based aggregation of the feature vectors of the neighbor points:

\begin{equation} \label{eq:gjsa}
   g_j^{sa} = \phi \left(\sum_{k=1}^{K} a_{j,k} \psi(f_{j,k}) \right)
\end{equation}

\begin{equation} \label{eq:ajk}
   a_{j,k} = \rho\left(\frac{\beta(g_j) \varphi(f_{j,k})^T}{\sqrt{d_{qk}}}\right)
\end{equation}

\noindent where $\rho$ is the softmax function, and $\phi$ is an MLP with ReLU activation function. In this attention-based feature aggregation scheme \citep{Vaswani2017AttYNeed}, $\beta(g_j)$ corresponds to the query vector, which is obtained through a linear transformation of $g_j$. The linear transformations of the feature vectors of the neighbor points, $\varphi(f_{j,k})$ and $\psi(f_{j,k})$ serve as key and value vectors, respectively. $d_{qk}$ is the dimension of the query and key vectors. The transformations applied to the feature vectors of the center point and neighbor points to obtain the query, key, and value vectors reduce their original dimension to half. The weights $a_{j,k}$ are determined through the dot product of the query vector $\beta(g_j)$ and the key vectors $\varphi(f_{j,k})$. In other words, the similarity between the feature vectors of the center point and the neighbor points are reflected to the weights $a_{j,k}$, which determine the contribution of the neighbor points to the aggregation. $g_j^{sa}$ is the output of $\phi$, which transforms the weighted sum of the value vectors and increases the dimension of the feature vector back to the original dimension of $g_j$.

The transformations involved in CSM-I are embedded in the functions $\gamma$, $\phi$, $\varphi$, $\beta$, and $\psi$ are learned through training. This scheme allows the output displacement vector $\Delta c_{j}$ be inferred through CSM, whose parameters are adjusted through minimization of a loss function designated for a particular task.

\subsubsection{CSM-II}

The main difference between CSM-I and CSM-II is in the determination of the weights $a_{j,k}$ used in Equation \ref{eq:gjsa}. In CSM-II variant, for updating the feature vector of the center point, positional relationships, $\delta_{jk}$, between query and key vectors are integrated into the computation of $a_{j,k}$:

\begin{equation} \label{eq:ajkII}
   a_{j,k} = \rho \left( \vartheta \left(concat\left[\delta_{jk}, d\left(\beta(g_j), \varphi(f_{j,k})\right)\right] \right) \right)
\end{equation}

\noindent where $\delta_{jk}$ is a transformed version of the relative position vector, $c_j - p_{j,k}$ through a linear transformation $\Theta: \mathbb{R}^3 \rightarrow \mathbb{R}^3$:

\begin{equation} \label{eq:b6_eq10}
   \delta_{jk} = \Theta(c_j - p_{j,k})
\end{equation}

$\rho$ in Equation \ref{eq:ajkII}  is the softmax function, and $\vartheta$ is a two-layer network with a ReLU activation function in between the layers. $d(.,.)$ defines the relation between the query and key vectors, $\beta(g_j)$ and $\varphi(f_{j,k})$. We considered various alternatives for the similarity measures of these query and key vectors, $d\left(\beta(g_j), \varphi(f_{j,k})\right)$, including subtraction (sub), summation (sum), concatenation (cat), and dot product (dot). This similarity is then integrated with positional relationship $\delta_{jk}$ via concatenation.

The rest of the operation of CSM-II is the same as CSM-I. The final displacement vector $\Delta c_{j}$ is obtained using Equations \ref{eq:deltaEjk} to \ref{eq:gjsa}.

\subsubsection{CSM-III}

In CSM-III, for determining the shift vector of the center point, relationships with close center points are incorporated in addition to the interactions between the particular center point and the points $p_{j,k}$ in its vicinity. The final displacement vector $\Delta c_{j}$ is obtained with Equations \ref{eq:deltaEjk} and \ref{eq:deltacj} as in CSM-I. However, the updated feature vector $\hat{g}_j$ of the center point is computed as:

\begin{equation} \label{eq:gjnonlocal}
   \hat{g}_j = g_j^{sa} + g_j^{saC}
\end{equation}

\noindent where $g_j^{sa}$ is obtained through the attention-based aggregation defined by Equations \ref{eq:gjsa} and \ref{eq:ajk}. $g_j^{saC}$ refers to the attention-based aggregation among the updated feature vectors belonging to the $U$ closest center points of the particular center point $c_j$. The feature vectors of the closest $U$ center points $g_{j,u}$, with $u=1,...U$, are updated as:

\begin{equation} \label{eq:b6_eq17_0}
   \bar{g}_{j,u} = g_{j,u} + g_{j,u}^{sa} 
\end{equation}

\noindent Then, $g_j^{saC}$ is computed as:

\begin{equation} \label{eq:b6_eq17}
   g_j^{saC} = \Tilde{\phi} \left(\sum_{u=1}^{U} b_{u} \Tilde{\psi}(\bar{g}_{{j,u}}) \right)
\end{equation}

\noindent with

\begin{equation} \label{eq:b6_eq16}
   b_{u} = \rho\left(\frac{\Tilde{\beta}(\bar{g}_{{j}}) \Tilde{\varphi}(\bar{g}_{{j,u}})^T}{\sqrt{d_{qk}}}\right)
\end{equation}

\noindent where $\rho$ is the softmax function, and $\Tilde{\phi}$ is a nonlinear transformation network consisting of an MLP with ReLU activation function. Here, $\Tilde{\beta}(\bar{g}_j)$ is the query vector. The linear transformations of the updated feature vectors of the $U$ center points, $\Tilde{\varphi}(\bar{g}_{{j,u}})$ and $\Tilde{\psi}(\bar{g}_{{j,u}})$ correspond to key and value vectors, respectively. With CSM-I and CSM-II, we only consider point interactions within the locality of radius $2r^L$ of each center point to shift it. With CSM-III, the spatial extent of point interactions is enlarged. The relative locations and feature similarities of other center points also contribute to the shift of a particular center point. 

\subsubsection{CSM-IV}

In CSM-IV, feature vectors of all center points $c_j$, $j=1,..,N_L$ are first updated using Equation \ref{eq:gjnonlocal}. While computing $g_j^{saC}$ component in Equation \ref{eq:b6_eq17}, all center points contribute rather than the nearest $U$ center points. Consequently, global interactions between a center point and all other center points covering the object are encoded in $g_j^{saC}$.   Likewise, the relative position vectors between $c_j$ and all other center points contribute to the shift. The displacement vector affected by another center point $c_l$ on $c_j$ is defined as:

\begin{equation} \label{eq:b6_eq19}
   \Delta E(j,l) = \theta (\hat{g}_j - \hat{g}_l)(c_j-c_l)
\end{equation}

\noindent where $\theta$ is a nonlinear transformation function that takes the difference of center points' features as input. 

The final shift vector is the average of $\Delta E(j,l)$ over all center points:

\begin{equation} \label{eq:b6_eq20}
   \Delta c_j = \frac{1}{N_L} \sum_{l=1}^{N_L} \Delta E(j,l)
\end{equation}

Among all versions of the Center Shift Module, CSM-IV considers point relationships most globally. It shifts sampled center locations by taking into account all center points covering the entire object.

\subsubsection{CSM-V}

CSM-V is fundamentally different from the other versions in that pairwise point interactions within $\mathcal{R}_j$ are employed. For each center point, $c_j$ weighted pairwise distances within its neighborhood $\mathcal{R}_j$ are formulated as 

\begin{equation} \label{eq:b6_eq21}
   \Delta E_j(k,l) = \theta (f_{j,k} - f_{j,l})(p_{j,k}-p_{j,l})
\end{equation}

The relative location vector between point $p_{j,k}$ and $p_{j,l}$ is weighted by the $3 \times 3$ output of the network $\theta$, which is a two-layered network with a ReLU activation function in between the layers. The pairwise feature differences are reduced to 64 and 9 dimensions in the first and second layer, respectively. The final displacement vector $\Delta c_{j}$ is calculated as  

\begin{equation} \label{eq:b6_eq23}
   \Delta c_j = \max_{k=1,...,K} \Delta G_{j,k} 
\end{equation}

\noindent with

\begin{equation} \label{eq:b6_eq22}
   \Delta G_{j,k} = \frac{1}{K} \sum_{l=1}^{K} \Delta E_j(k,l)
\end{equation}

Another important distinction of CSM-V from the first four variations is that feature update through attention-based aggregation is not employed prior to calculation of feature interactions.

\subsection{Radius update module}
\label{sec:RUM}

RUM is responsible to update the radius of the neighborhood of each center point $c_j$. The convention is setting the radius $r^{L}$ constant for all center points at a particular layer. We propose to update the radius for each center point by $\Delta r_{j}$ produced by RUM, through Equation \ref{eq:radius}. In this way, the size of the receptive field is defined in an adaptive way for each center point. 

For the sake of simplicity, the superscript $L$ is dropped. The points from the previous layer within $2r$ distance of a center point $c_j$ are gathered randomly. Let the feature vectors of these points be denoted as $f_{j,s}$, $s=1,..,S$, where $S$ is the number of points in the neighborhood of radius $2r$. Let the feature vector of the center point be denoted as $g_j$. We propose two variations of RUM for inferring $\Delta r_{j}$ from these points.

\subsubsection{RUM-I}

First, the difference between the feature vectors of the center point and each neighbor point is transformed through the nonlinear function $\zeta$: 

\begin{equation} \label{eq:b6_eq24}
   e_{j,s} = \zeta(g_j - f_{j,s})
\end{equation}

\noindent where $\zeta$ is an MLP with ReLU activation function. The spherical region of radius $2r$ surrounding $c_j$ is partitioned into $T$ concentric spheres, $\mathcal{B}_{j,t}$, each with radius $\frac{t}{T}2r$. The number of points in $\mathcal{B}_{j,t}$ is denoted as $S_{j,t}$. Transformed feature differences of the points in each region is aggregated through either averaging 

\begin{equation} \label{eq:Rcum}
   R_{j,t} = \frac{1}{S_{j,t}}\sum_{s \in \mathcal{B}_{j,t}} e_{j,s}
\end{equation}

\noindent or maxpooling

\begin{equation} \label{eq:Rmax}
   R_{j,t} = \max_{s \in \mathcal{B}_{j,t}} e_{j,s} 
\end{equation}.

In the experiments, we considered and compared both alternatives for  feature aggregation as RUM-I (cum) for Equation \ref{eq:Rcum} and RUM-I (max) for Equation \ref{eq:Rmax}. The aggregated features $R_{j,t}$ are arranged in a matrix $\bar{R}_j$ following the order $t=1,...,T$ (i.e., from smaller to larger spheres). Since $\bar{R}_j$ is invariant to permutation of point indices, we apply an MLP with activation function $tanh$ to produce a scalar value, $\Delta r_j$, which is the output of RUM-I. The activation function $tanh$ is used to limit the magnitude of $\Delta r_j$.

\subsubsection{RUM-II}

In RUM-II, the neighborhood of each center point $c_j$ is organized by the same manner as in RUM-I. The features aggregated for each concentric sphere $t$ are updated as

\begin{equation} \label{eq:b6_eq30}
   \hat{R}_{j,t} = R_{j,t}+ R_{j,t}^{sa} 
\end{equation}

To obtain $R_{j,t}^{sa}$, an attention-based transformation is applied to the features $R_{j,t}$:

\begin{equation} \label{eq:b6_eq29}
   R_{j,t}^{sa} = \sum_{v=1}^{T}  a_{t,v} \hat\psi(R_{j,v})
\end{equation}

\noindent with weights $a_{t,v}$ computed as

\begin{equation} \label{eq:b6_eq28}
   a_{t,v} = \rho\left(\frac{\hat\beta(R_{j,t}) \hat\varphi(R_{j,v})^T}{\sqrt{d_{qk}}}\right)
\end{equation}

Here, $\rho$ is the softmax function, $\hat\beta(R_{j,t})$ is a linear transformation of $R_{j,t}$, and corresponds to the query vector. The linear transformations of $R_{j,v}$, $v=1,...,T$ are designated as the key $\hat\varphi(R_{j,v})$ and value $\hat\psi(R_{j,v})$ vectors, with dimension $d_{qk}$. For RUM-II, these linear transformations do not involve dimension reduction.

As in RUM-I, the features $\hat{R}_{j,t}$ are arranged to form the matrix $\bar{R}_j$, and $\Delta r_j$ is computed by an MLP with $tanh$ activation function. For both RUM-I and RUM-II, we set $T=4$ in our experiments.

\subsection{Loss function}

In 3D point-based classification architectures, usually, cross-entropy is used as the loss function. We introduced additional terms, $\mathcal{L}_{csm}^L$, and $\mathcal{L}_{rum}^L$, to the loss function in order to limit the shift generated by CSM and the radius update introduced by RUM:

\begin{equation} \label{eq:loss}
   \mathcal{L} = \mathcal{L}_{ce} + \alpha_1 \sum_{L}{\mathcal{L}_{csm}^L} + \alpha_2\sum_{L}{\mathcal{L}_{rum}^L}
\end{equation}

\noindent where $\mathcal{L}_{ce}$ is the cross-entropy function and $\alpha_1$ and $\alpha_2$ are constant weights, which are set to $0.01$ in our experiments. $\mathcal{L}_{ce}$ ensures that the network learns the parameters of both feature extraction layers and CSM and RUM modules via maximizing the classification performance on the training data. With $\mathcal{L}_{csm}^L$, we introduce a regularization term that penalizes large shifts of the center points computed at layer $L$. $\mathcal{L}_{csm}^L$ is composed of two terms:

\begin{equation} \label{eq:b6_eq33}
   \mathcal{L}_{csm}^L = \mathcal{L}_{fit}^L + \mathcal{L}_{range}^L
\end{equation}

\begin{equation} \label{eq:b6_eq34}
   \mathcal{L}_{fit}^L = \frac{1}{N_L} \sum_{j=1}^{N_L}  \lVert \hat{c}_{j}^L - \Tilde{p}_{j}^L \rVert
\end{equation}

\begin{equation} \label{eq:b6_eq35}
   \mathcal{L}_{range}^L =  \frac{1}{N_L} \sum_{j=1}^{N_L} \max (0, \lVert \Delta c_{j}^L \rVert - r^L) 
\end{equation}

\noindent where $\Tilde{p}_{j}^L$ is the closest point to the updated center point $\hat{c}_{j}^L$ from the set $\mathcal{P}^{L-1}$. $\mathcal{L}_{fit}^L$ forces the updated center points $\hat{c}_{j}^L$ to remain close to the object surface, while $\mathcal{L}_{range}^L$ penalizes the excess in the magnitude of the shift vector $\Delta c_{j}^L$ in relation to $r^L$, the initial radius of local regions at a particular layer $L$. 

The other term, $\mathcal{L}_{rum}^L$, limits the magnitude of the radius update $\Delta r_{j}^L$:

\begin{equation} \label{eq:b6_eq36}
    \begin{split}
       \mathcal{L}_{rum}^L =\frac{1}{N_L} \sum_{j=1}^{N_L} \bigl( \lvert \min (0, r^L + \Delta r_{j}^L) \rvert +  \\ \max (0, \Delta r_{j}^L - r^L) \bigl)
    \end{split}
\end{equation}

\noindent In this way, $\Delta r_{j}^L$ is encouraged to remain in the interval $[-r^L,r^L]$.

\begin{figure}[h]
    \centering
    \begin{subfigure}[b]{1.0\textwidth}
         \centering
         \includegraphics[width=\textwidth]{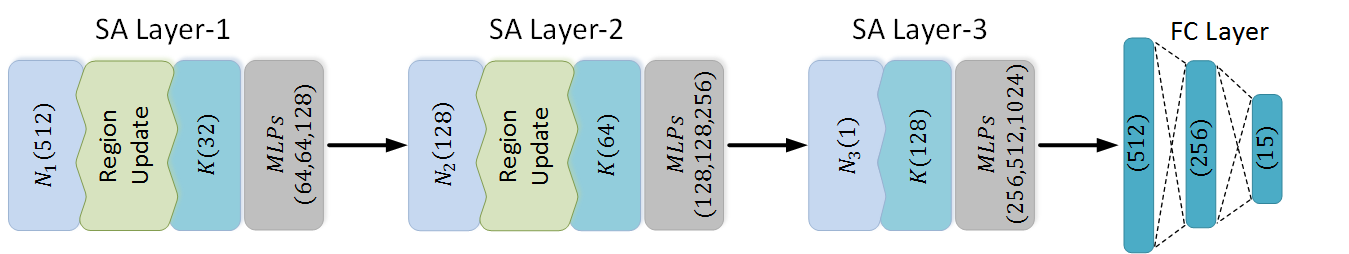}
         \caption{}
         \label{fig:poitnet2_network}
    \end{subfigure}
    \begin{subfigure}[b]{1.0\textwidth}
         \centering
         \includegraphics[width=\textwidth]{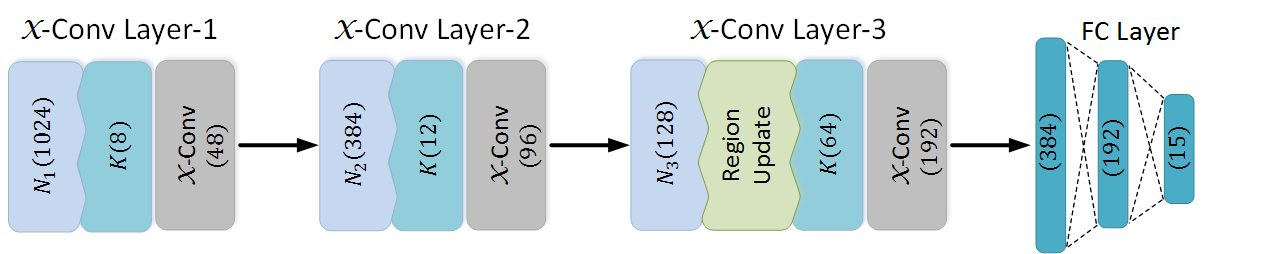}
         \caption{}
         \label{fig:pointcnn_network}
    \end{subfigure}
    \begin{subfigure}[b]{0.8\textwidth}
         \centering
         \includegraphics[width=\textwidth]{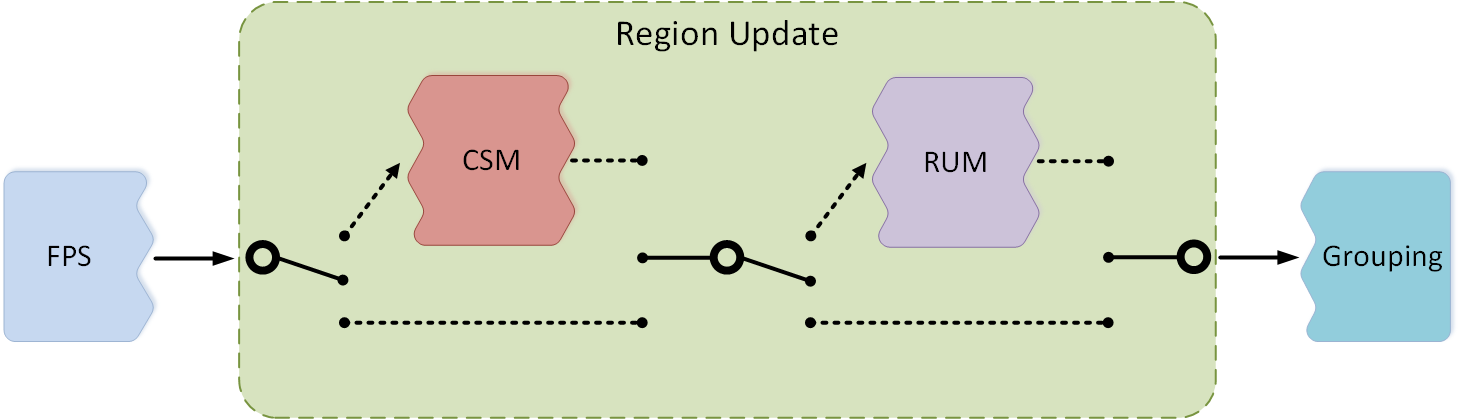}
         \caption{}
         \label{fig:poitnet2_modules}
     \end{subfigure}
     \caption{The integration of proposed modules to PointNet++ and PointCNN frameworks: a-) PointNet++ network for the classification problem. $r^L$ for $L = 1$ (first layer) is set to be 0.2. $r^L$ for $L = 2$ (second layer) is 0.4. b-) PointCNN network for the classification problem. $r^L$ for $L = 3$ (third layer) is set to be 0.4. c-) adaptive local region inference framework.}
     \label{fig:pointnet2}
\end{figure}

\subsection{Integration to 3D object classification networks}

The proposed local region-learning modules, CSM and RUM can be integrated into any hierarchical deep learning framework consisting of sampling and grouping stages, where group centers and the sizes of the regions occupied by the groups are needed to be determined. We selected two 3D object classification architectures to demonstrate the performance gain by utilizing CSM and RUM: 1) PointNet++ classification network \citep{Qi2017PointNet2}, which was employed in a wide range of applications \citep{Yang_2020_CVPR, Briechle2020trees, ZHOU2022powder, Ma2022Lidar}; and 2) PointCNN \citep{Li2018PointCNN} as a state-of-the-art network, which yielded the highest performance for OBJ\_BG set of ScanObjectNN data set \citep{Uy2019ScanObjectNN}.

\subsection{Integration to PointNet++}

The basic structure of the classification network architecture of PointNet++ \citep{Qi2017PointNet2} is shown in Figure \ref{fig:pointnet2}a. The global descriptors representing each object are extracted by passing its point cloud representation through three Set Abstraction (SA) layers. The category scores are obtained by applying fully connected layers to the global descriptor. 

At each SA layer, the center points $c_j^L$ of $N_L$ local regions are determined with the FPS algorithm. CSM and RUM can be integrated to either or both of the first two layers as shown in Figure \ref{fig:pointnet2}c. If CSM is "ON" at a particular layer, the center points are shifted by $\Delta c_j^L$ as in Equation \ref{eq:hatcj}, otherwise they are not altered. If RUM is "OFF", a local region with radius $r^L$, which is kept constant for layer $L$, is established around each center point. If RUM is "ON", then the radius is updated using Equation \ref{eq:radius}, and a local region of radius $r_j^L$ is formed around center point $c_j^L$. Grouping is performed by randomly selecting $K$ points in the local regions. After the features of the $K$ points are mapped to higher dimensions independently with MLP layers, the features of the local region centers are calculated by taking the maximum among the feature channels of grouped points.

The third abstraction layer is responsible for computing the global descriptor for the entire point cloud of the object. At this layer, a single center point is defined, with its "local region" covering all the center points conveyed from the previous layer. 

\subsubsection{Integration to PointCNN}
The hierarchical classification framework of PointCNN is illustrated in Figure \ref{fig:pointnet2}b. PointCNN leverages a specialized convolution operator known as $\mathcal{X}$-Conv. This operator serves a dual purpose: firstly, it permutes the points within a given set into a canonical order and secondly, it weights the input features associated with the points in local regions. The point ordering and weight assignment are performed through the utilization of a $K{\times}K$ transformation, which is learned through training of multi-layer perceptrons. 

The PointCNN classification network contains three $\mathcal{X}$-Conv layers, which systematically embed input features (point coordinates at the first layer) into a higher-dimensional feature space while reducing the number of representative points. The network also includes a classification layer designed to obtain category scores, similar to the approach used in PointNet++.


The PointCNN framework involves selecting representative points via utilization of random sampling or farthest point sampling algorithm. The constitution of local regions centered around each of these representative points is achieved through performing of a dilated nearest neighbor search algorithm.

Upon observing that CSM and RUM modules are most effective when they are applied to the last sampling stage of PointNet++ (see Section \ref{sec:resultsPointNetPlus}), we decided to integrate CSM and RUM only to the last $\mathcal{X}$-Conv layer of PointCNN. Initial representative points are selected through farthest point sampling. Instead of using a dilated nearest neighbor search algorithm to select $K$ points within a neighborhood, we randomly sampled $K$ points within spherical regions to be transformed by the $\mathcal{X}$-Conv operation.

\section{Results}
\label{section:result}

We conducted various experiments to observe the effect of CSM and RUM on the performance of PointNet++ and PointCNN. The experiments were performed on the real-world classification data set ScanObjectNN \citep{Uy2019ScanObjectNN} and on the ShapeNet data set \citep{Chang2015shapenet}, which contains CAD models of objects.

First, variations of only CSM or only RUM were integrated to the set abstraction layers of PointNet++ to compare their performance on ScanObjecNN. The modules were either integrated to the first layer, or to the second, or to both. Then, various combinations of versions of CSM and RUM were inserted together to the second layer of PointNet++. Similar experiments were performed with PointCNN on ScanObjectNN. We integrated the modules only to the third $\mathcal{X}$-Conv layer of PointCNN. Finally, we tested selected versions of the modules integrated to PointNet++ and PointCNN on ShapeNet data set.

\begin{table*}[h]
\caption{Results of PointNet++ with CSM on ScanObjectNN data set.}
\centering
\begin{tabular}{lcccccccccc} \hline 
 & & &\multicolumn{4}{c}{OBJ\_ONLY} &\multicolumn{4}{c}{OBJ\_BG} \\  \cmidrule(lr){4-7} \cmidrule(lr){8-11}
\textbf{Method} &1\textsuperscript{st} &2\textsuperscript{nd} &$Acc$ &+/- &$MAcc$ &+/- &$Acc$ &+/- &$MAcc$ &+/- \\ \hline
PointNet++ & & &84.33 &- &82.1 &- &82.3 &- &79.9 &-\\ \hline 
\multirow{3}{6em} {CSM-I} &\checkmark & &85.71 &\textbf{$+$1.38} &83.62 &\textbf{$+$1.52} &86.75 &\textbf{$+$4.45} &83.57 &\textbf{+3.67} \\ & &\checkmark &87.44 &\textbf{$+$3.11} &84.89 &\textbf{$+$2.79} &86.75 &\textbf{$+$4.45} &85.15 &\textbf{$+$5.25}  \\ &\checkmark &\checkmark &85.03 &\textbf{$+$0.70} &82.59 &\textbf{$+$0.49} &86.06 &\textbf{$+$3.76} &83.08 &\textbf{$+$3.18} \\ \hline
\multirow{3}{6em} {CSM-II (sub)} &\checkmark & &85.20 &\textbf{$+$0.87} &83.84 &\textbf{$+$1.74} &86.06 &\textbf{$+$3.76} &81.90 &\textbf{$+$2.00} \\ & &\checkmark &87.26 &\textbf{$+$2.93} &85.31 &\textbf{$+$3.21} &86.75 &\textbf{$+$4.45} &84.61 &\textbf{$+$4.71} \\ &\checkmark &\checkmark &84.38 &\textbf{$+$0.05} &83.36 &\textbf{$+$1.26} &86.58 &\textbf{$+$4.28} &82.75 &\textbf{$+$2.85} \\ \hline 
\multirow{3}{6em} {CSM-II (sum)} &\checkmark & &86.06 &\textbf{$+$1.73} &84.92 &\textbf{$+$2.82} &87.09 &\textbf{$+$4.79} &83.96 &\textbf{$+$4.06} \\ & &\checkmark &86.92 &\textbf{$+$2.59} &85.32 &\textbf{$+$3.22} &88.47 &\textbf{$+$6.17} &86.07 &\textbf{$+$6.17} \\ &\checkmark &\checkmark &86.75 &\textbf{$+$2.42} &85.23 &\textbf{$+$3.13} &87.09 &\textbf{$+$4.79} &85.13 &\textbf{$+$5.23} \\ \hline 
\multirow{3}{6em} {CSM-II (cat)} &\checkmark & &84.51 &\textbf{$+$0.18} &81.98 &\textbf{$+$0.12} &86.06 &\textbf{$+$3.76} &82.94 &\textbf{$+$3.04} \\ & &\checkmark &85.37 &\textbf{$+$1.04} &83.93 &\textbf{$+$1.83} &86.06 &\textbf{$+$3.76} &83.60 &\textbf{$+$3.70}  \\ &\checkmark &\checkmark &86.23 &\textbf{$+$1.90} &83.28 &\textbf{$+$1.18} &86.58 &\textbf{$+$4.28} &83.31 &\textbf{$+$3.41} \\ \hline 
\multirow{3}{6em} {CSM-II (dot)} &\checkmark & &85.03 &\textbf{$+$0.70} &81.93 &-0.17 &85.37 &\textbf{$+$3.07} &82.28 &\textbf{+2.38} \\ & &\checkmark &86.23 &\textbf{$+$1.90} &85.46 &\textbf{$+$3.36} &86.92 &\textbf{$+$4.62} &85.13 &\textbf{$+$5.23} \\ &\checkmark &\checkmark &87.44 &\textbf{$+$3.11} &86.02 &\textbf{$+$3.92} &85.89 &\textbf{$+$3.59} &82.88 &\textbf{$+$2.98} \\ \hline 
\multirow{3}{10em} {CSM-II (hadamard)} &\checkmark & &85.20 &\textbf{$+$0.87} &83.81 &\textbf{$+$1.71} &86.58 &\textbf{$+$4.28} &84.27 &\textbf{$+$4.37} \\ & &\checkmark &86.06 &\textbf{$+$1.73} &84.34 &\textbf{$+$2.24} &86.23 &\textbf{$+$3.93} &83.36 &\textbf{$+$3.46} \\ &\checkmark &\checkmark &85.54 &\textbf{$+$1.21} &84.00 &\textbf{$+$1.90} &87.26 &\textbf{$+$4.96} &84.27 &\textbf{$+$4.37} \\ \hline 
\multirow{3}{6em} {CSM-III} &\checkmark & &85.89 &\textbf{$+$1.56} &83.99 &\textbf{$+$1.89} &87.78 &\textbf{$+$5.48} &85.26 &\textbf{$+$5.36} \\ & &\checkmark &88.81 &\textbf{$+$4.48} &86.74 &\textbf{+4.64} &87.09 &\textbf{$+$4.79} &84.90 &\textbf{$+$5.00} \\ &\checkmark &\checkmark &85.54 &\textbf{$+$1.21} &83.43 &\textbf{$+$1.33} &85.71 &\textbf{$+$3.41} &82.94 &\textbf{$+$3.04} \\ \hline 
\multirow{3}{6em} {CSM-IV} &\checkmark & &84.34 &\textbf{$+$0.01} &81.99 &$-$0.11 &86.92 &\textbf{$+$4.62} &84.56 &\textbf{$+$4.66} \\ & &\checkmark &86.06 &\textbf{$+$1.73} &84.35 &\textbf{$+$2.25} &87.61 &\textbf{$+$5.31} &83.97 &\textbf{$+$4.07} \\ &\checkmark &\checkmark &84.85 &\textbf{$+$0.52} &83.66 &\textbf{$+$1.56} &86.40 &\textbf{$+$4.10} &83.98 &\textbf{$+$4.08}\\ \hline 
\multirow{3}{6em} {CSM-V} &\checkmark & &85.37 &\textbf{$+$1.04} &83.15 &\textbf{$+$1.05} &86.92 &\textbf{$+$4.62} &82.94 &\textbf{$+$3.04} \\ & &\checkmark &87.09 &\textbf{$+$2.76} &84.68 &\textbf{$+$2.58} &88.47 &\textbf{$+$6.17} &85.11 &\textbf{$+$5.21} \\ &\checkmark &\checkmark &83.99 &$-$0.34 &81.19 &$-$0.91 &85.03 &\textbf{$+$2.73} &81.86 &\textbf{$+$1.96} \\ 

\hline 
\end{tabular}
\label{tab:tablecsm}
\end{table*}

\begin{figure}[h]
     \centering
     \begin{subfigure}[b]{0.49\textwidth}
         \centering
         \includegraphics[width=\textwidth]{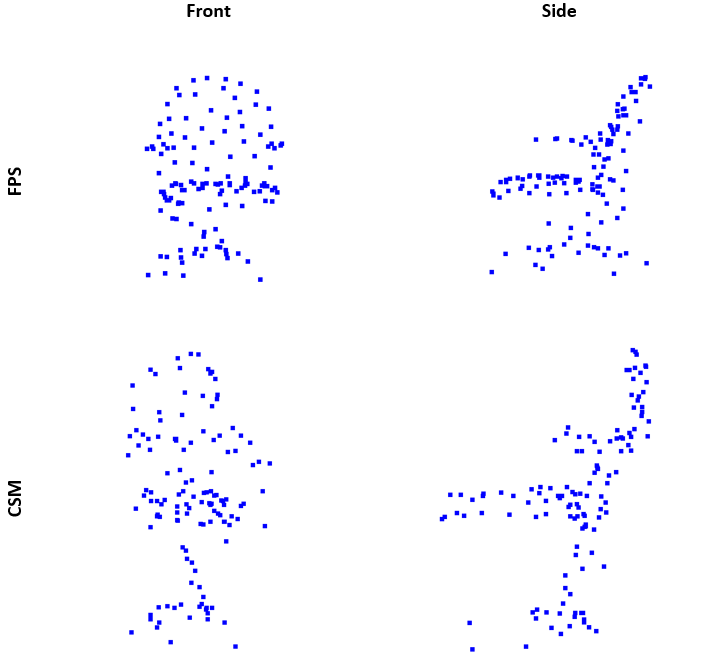}
         \caption{}
         \label{fig:csm_a}
     \end{subfigure}
     \begin{subfigure}[b]{0.47\textwidth}
         \centering
         \includegraphics[width=\textwidth]{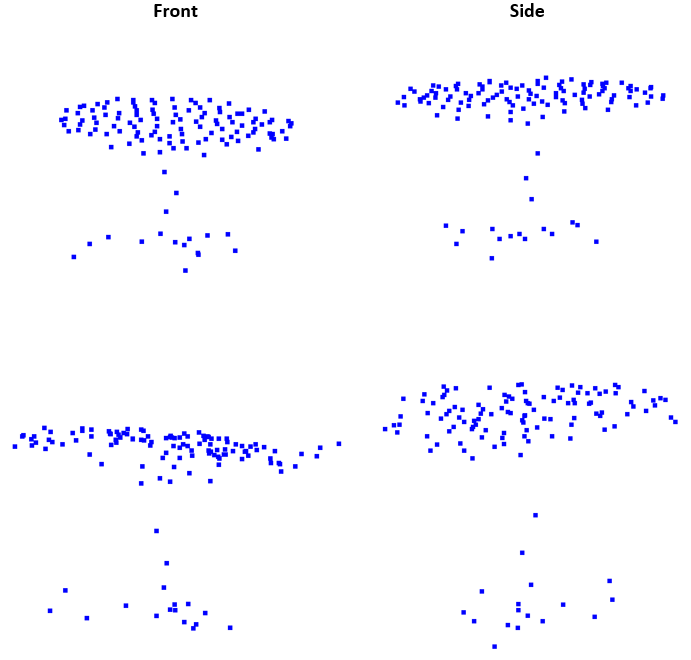}
         \caption{}
         \label{fig:csm_c}
     \end{subfigure}
     \begin{subfigure}[b]{0.49\textwidth}         
         \centering
         \includegraphics[width=\textwidth]{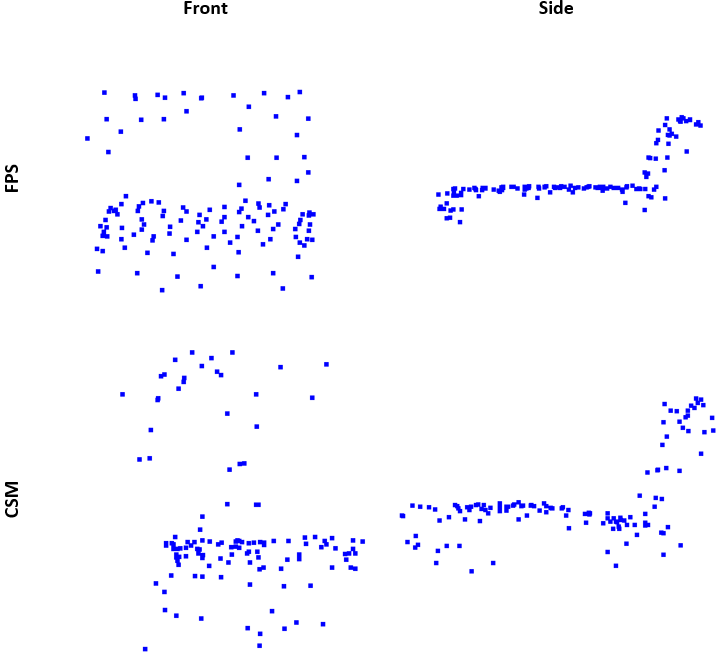}
         \caption{}
         \label{fig:csm_b}
     \end{subfigure}

     \caption{Shifted local center points for sample objects from OBJ\_ONLY. The objects belong to the chair (a),  table (b), and bed (c) categories. CSM-III is integrated only to the second layer of PointNet++. 
     \label{fig:csmvisual}}
\end{figure}

\subsection{Data Sets}
\label{subsection:dataset}

\subsubsection{ScanObjectNN}
\label{subsection:scanobjectNN}

The ScanObjectNN \citep{Uy2019ScanObjectNN} data set consists of real-world point cloud object data collected from the public 3D indoor scene data sets SceneNN \citep{Nguyen2017SceneNN} and ScanNet \citep{Dai2017Scannet}. 700 unique scenes of SceneNN and ScanNet mesh indoor scans were selected, and from these scenes, 2902 objects were cropped and manually filtered. These objects are categorized into 15 common categories (bag, bed, bin, box, cabinet, chair, desk, display, door, pillow, shelf, sink, sofa, table, and toilet). Each point cloud is represented with local and global coordinates, surface normals, color attributes, and semantic labels. Different variants of each object have been generated to add  levels of difficulty to explore the robustness of classification algorithms. We experimented with two variants referred to as OBJ\_ONLY and OBJ\_BG. The first variant, OBJ\_ONLY, includes objects neatly delineated from the background. The second variant, OBJ\_BG, corresponds to an axis-aligned bounding box around the object, and contains parts of the background and nearby objects as well.

We used the training and test split and the parameters as specified in \citep{Uy2019ScanObjectNN}. For all experiments, each point cloud was randomly sampled to 1024 points, centered at zero, and scaled to fit in the unit sphere. As input features, only the local coordinates ($x, y, z$) of the points were used.

\subsubsection{ShapeNet}
\label{subsection:shapenet}

We conducted additional experiments on the ShapeNet dataset \citep{Chang2015shapenet}, which consists of 16,881 3D polygonal CAD models belonging to 16 categories  (airplane, bag, cap, car, chair, earphone, guitar, knife, lamp, laptop, motorbike, mug, pistol, rocket, skateboard, and table). We used the point clouds that had been extracted from the polygon models of ShapeNet and been preprocessed by the creators of PointNet++ \citep{Qi2017PointNet2}. Each point cloud model includes local point coordinates and surface normals as attributes and semantic labels.

We followed the official training and test splits of ShapeNet benchmark \citep{Chang2015shapenet}. Similar to the experiments on ScanObjectNN data set, each point cloud in ShapeNet was randomly sampled to 1024 points and normalized in scale. In the experiments, surface normal attributes are dropped and only local point coordinates are used.


\subsection{Experimental results}
\label{subsection:experiments}

\subsubsection{Results with CSM and RUM integrated to PointNet++}
\label{sec:resultsPointNetPlus}

With PointNet++, three types of experiments were performed for each variation of CSM and RUM: 1) the module is applied only to the first layer (1\textsuperscript{st}), 2) the module is applied only to the second layer (2\textsuperscript{nd}), and 3) the module is applied to both the first and second layers (1\textsuperscript{st}-2\textsuperscript{nd}). The results are compared in terms of overall accuracy ($Acc$), which is equal to the ratio of the number of correctly classified objects to the total number of objects in the test set. We also give the mean of the accuracy ($MAcc$), which corresponds to the class-based accuracy values averaged over all categories.

Table \ref{tab:tablecsm} gives the contribution of variations of CSM to the classification performance of PointNet++ on the ScanObjectNN data sets OBJ\_ONLY and OBJ\_BG. In the first row, the classification accuracy of PointNet++ is provided as reported in \citep{Uy2019ScanObjectNN}. All variations of CSM, except for CSM-IV applied only to the first layer, and CSM-V applied to both layers, increased the performance of PointNet++. However, for most of the variations, applying CSM only to the second layer resulted in a higher increase in performance than applying it to only the first layer or to both. At the very local level, keeping the center points selected via FPS gives close results to those obtained with moving the center points with CSM. As the region size grows in the second layer, semantic content aggregated at center points gains prominence for classification. CSM module displaces these center points in accord with their contribution to the minimization of the task loss.

\begin{figure}[h]
     \centering
     \begin{subfigure}[b]{0.49\textwidth}
         \centering
         \includegraphics[width=\textwidth]{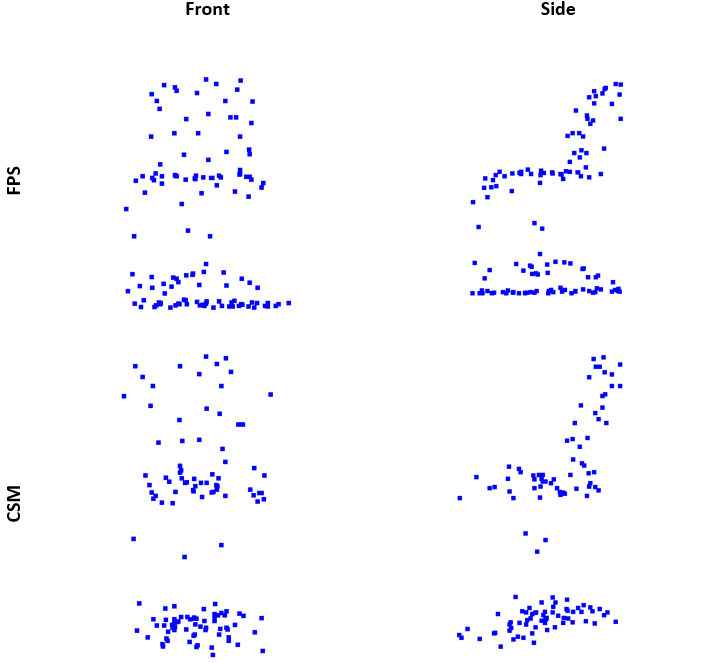}
         \caption{}
         \label{fig:csm_bg_a}
     \end{subfigure}
     \begin{subfigure}[b]{0.47\textwidth}
         \centering
         \includegraphics[width=\textwidth]{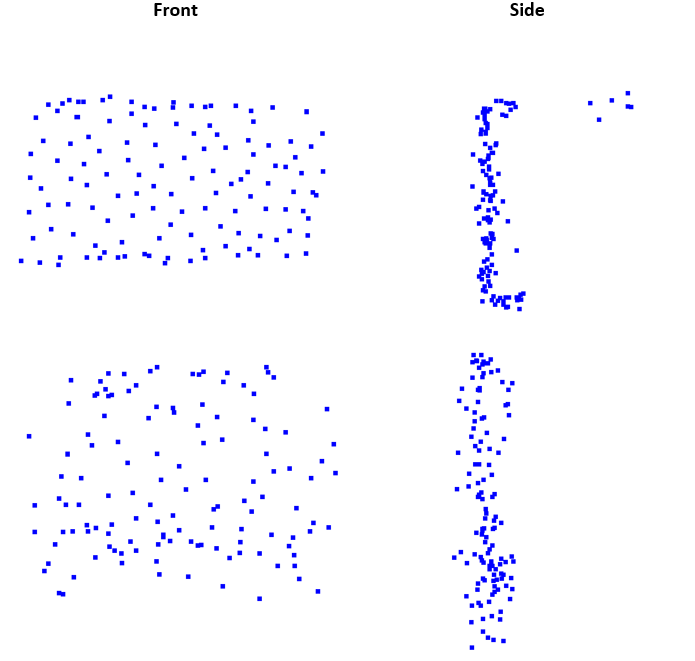}
         \caption{}
         \label{fig:csm_bg_b}
     \end{subfigure}

     \caption{
     Shifted local center points for sample objects from OBJ\_BG. The objects belong to the chair (a) and display (b) categories. CSM-II-(sum) is integrated only to the second layer of PointNet++.}
     \label{fig:visualcsmbg}
\end{figure}

\begin{table*}[h]
\caption{Results of PointNet++ with RUM on ScanObjectNN data set.}
\centering
\begin{tabular}{lcccccccccc} \hline 
 & & &\multicolumn{4}{c}{OBJ\_ONLY} &\multicolumn{4}{c}{OBJ\_BG} \\  \cmidrule(lr){4-7} \cmidrule(lr){8-11}
\textbf{Method} &1\textsuperscript{st} &2\textsuperscript{nd} &$Acc$ &+/- &$MAcc$  &+/- &$Acc$  &+/- &$MAcc$  &+/-\\ \hline
PointNet++ & & &84.33 &- &82.1 &- &82.3 &- &79.9 &- \\ \hline 
\multirow{3}{8em} {RUM-I (cum)} &\checkmark & &82.79 &$-$1.54 &80.52 &$-$1.58 &83.99 &\textbf{$+$1.69} &81.60 &\textbf{$+$1.70} \\ & &\checkmark &87.09 &\textbf{$+$2.76} &85.45 &\textbf{$+$3.35} &86.92 &\textbf{$+$4.62} &85.17 &\textbf{$+$5.27} \\ &\checkmark &\checkmark &86.23 &\textbf{$+$1.90} &84.30 &\textbf{$+$2.20} &85.20 &\textbf{$+$2.90} &82.17 &\textbf{$+$2.27} \\ \hline 
\multirow{3}{8em} {RUM-I (max)} &\checkmark & &83.48 &$-$0.85 &81.40 &$-$0.7 &85.54 &\textbf{$+$3.24} &81.88 &\textbf{$+$1.98} \\ & &\checkmark &87.95 &\textbf{$+$3.62} &85.66 &\textbf{$+$3.56} &87.44 &\textbf{$+$5.14} &85.81  &\textbf{$+$5.91}\\ &\checkmark &\checkmark &84.85 &\textbf{$+$0.52} &82.81 &\textbf{$+$0.71} &86.40 &\textbf{$+$4.10} &84.00 &\textbf{$+$4.10} \\ \hline 
\multirow{3}{8em} {RUM-II (cum)} &\checkmark & &83.99 &$-$0.34 &82.22 &\textbf{$+$0.12} &85.37 &\textbf{$+$3.07} &82.49 &\textbf{$+$2.59} \\ & &\checkmark &87.44 &\textbf{$+$3.11} &86.13 &\textbf{$+$4.03} &88.30 &\textbf{$+$6.00} &86.31  &\textbf{$+$6.41} \\ &\checkmark &\checkmark &85.54 &\textbf{$+$1.21} &83.58 &\textbf{$+$1.48} &86.40 &\textbf{$+$4.10} &83.89 &\textbf{$+$3.99} \\ \hline 
\multirow{3}{8em} {RUM-II (max)} &\checkmark & &84.68 &\textbf{$+$0.35} &81.83 &$-$0.27 &85.71 &\textbf{$+$3.41} &81.86 &\textbf{$+$1.96} \\ & &\checkmark &85.71 &\textbf{$+$1.38} &83.58 &\textbf{$+$1.48} &87.44 &\textbf{$+$5.14} &85.87  &\textbf{$+$5.97} \\ &\checkmark &\checkmark &86.58 &\textbf{$+$2.25} &84.87 &\textbf{$+$2.77} &87.78 &\textbf{$+$5.48} &85.34 &\textbf{$+$5.44} \\ 

\hline 
\end{tabular}
\label{tab:tablerum}
\end{table*}

We provide visual results of the effect of CSM for sample objects belonging to various categories from the OBJ\_ONLY set in Figure \ref{fig:csmvisual}. Here, CSM-III was applied only to the second layer of PointNet++. In the first row for each object, center points sampled by FPS algorithm are given from front and side views. In the second row, center points shifted by CSM are shown. We can observe that CSM leaves some portions of the object with relatively few center points, while populating other parts with more representative points. We can view this tendency as the network's response to emphasize certain regions that are salient within a category. For example, as seen in Figure \ref{fig:csmvisual}, the center points of objects in the chair, table, and bed categories at the flat portions were kept in the same plane, however points at legs were widely displaced to counter intra-class variability. Legs are the portions that cause the most intra-class shape variation in categories such as chair and table. The salient part is the flat portion and its relative size with respect to the other parts. 

For all variations of CSM, application to the second layer increases the classification performance of PointNet++. For OBJ\_ONLY data set, the highest increase in $MAcc$ is +4.64\%, achieved by CSM-III.

The increase in classification accuracy with integration of CSM is more pronounced for the OBJ\_BG data, reaching +6.17 of $MAcc$ with CSM-II (sum). Two examples from the OBJ\_BG data are given in Figure \ref{fig:visualcsmbg} to observe the displacements of center points affected by CSM-II (sum). The main difference of the chair object in Figure \ref{fig:visualcsmbg}a as compared to the chair object in Figure \ref{fig:csmvisual}a is the presence of the ground points, which do not belong to the object but to the background. However, the ground points still provide context for the distinction of the chair category. Here, again, we observe that the non-salient chair points at legs that show greater within class variability tend to move to populate the salient portions such as the seat, and the ground in the case of OBJ\_BG. The points of the back of the chair moved widely. We conjecture that these points tend to be distributed to even out the variability of the back of the chairs in the training data set. We observe a similar effect for the display object shown in Figure \ref{fig:visualcsmbg}b, where the center points remained on the plane, however the distribution on the plane is altered to fit a certain aspect ratio. We can also notice that the center points behind the display are shifted towards the display to suppress the contribution of the points belonging to the irrelevant background.

Table \ref{tab:tablerum} gives the increase in the classification performance of PointNet++ by integrating variations of RUM. For OBJ\_ONLY data, applying RUM to the first layer did not contribute to an increase in performance; on the contrary, caused a drop with most variations. However, similar to the case with CSM, applying RUM only to the second layer resulted in higher increase in performance than applying it to both layers. The strategy of keeping the region size fixed for extracting low level surface features at the first layer while varying region size in the second layer where partial semantic information comes into prominence proves to be effective.

\begin{figure}[h]
     \centering
     \begin{subfigure}[b]{0.85\textwidth}
         \centering
         \includegraphics[width=\textwidth]{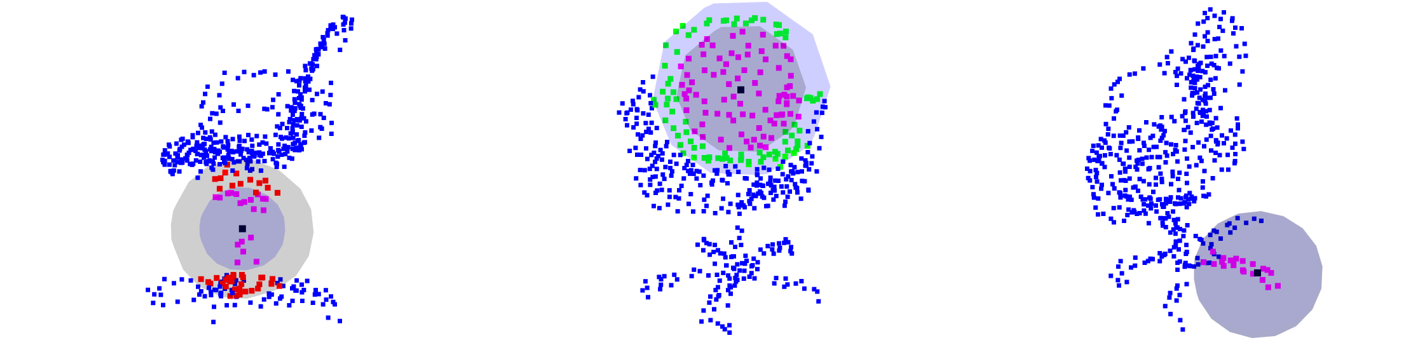}
         \caption{}
         \label{fig:rum_a}
     \end{subfigure}
     \begin{subfigure}[b]{0.85\textwidth}
         \centering
         \includegraphics[width=\textwidth]{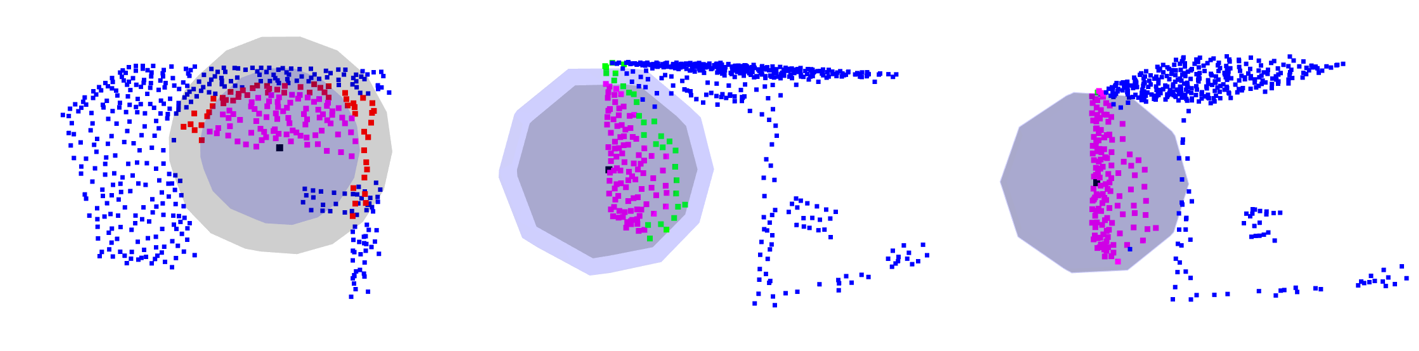}
         \caption{}
         \label{fig:rum_b}
     \end{subfigure}
     
     \caption{Updated region radii for sample objects from OBJ\_ONLY through RUM-I-(max) applied only to the second layer. The objects belong to the chair(a) and desk(b) categories. Black points represent the centers of the updated local regions. The red points are the points excluded from the region when the radius is decreased by RUM. The  green points indicate the points included to the region when the radius is increased. Points in magenta remained in the region.}
     \label{fig:rumvisual}
\end{figure}

\begin{table*}[!h]
\caption{Results of PointNet++ with merged combinations of CSM and RUM on ScanObjectNN data set.}
\centering
\begin{tabular}{lccccccccc} \hline 
 & &\multicolumn{4}{c}{OBJ\_ONLY} &\multicolumn{4}{c}{OBJ\_BG}\\ \cmidrule(lr){3-6} \cmidrule(lr){7-10}
\textbf{Method} &2\textsuperscript{nd} &$Acc$ &+/- &$MAcc$ &+/- &$Acc$ &+/- &$MAcc$ &+/- \\ \hline
PointNet++ & &84.33 &- &82.1 &- &82.3 &- &79.9 &- \\ \hline 
CSM-I  &\checkmark &88.47 &\textbf{$+$4.14} &87.35 &\textbf{$+$5.25}  &89.16 &\textbf{$+$6.86} &86.65 &\textbf{$+$6.75} \\ RUM-I (max) & & &  \\ \hline 
CSM-I &\checkmark &87.44 &\textbf{$+$3.11} &85.69 &\textbf{$+$3.59} &87.78 &\textbf{$+$5.48} &85.81 &\textbf{$+$5.91} \\ RUM-II (cum) & & &   \\ \hline
CSM-II (sub) &\checkmark  &88.30 &\textbf{$+$3.97} &87.23 &\textbf{$+$5.13}  &87.78 &\textbf{$+$5.48} &84.76 &\textbf{$+$4.86} \\ RUM-I (max) & & &   \\ \hline
CSM-II (sub) &\checkmark  &87.09 &\textbf{$+$2.76} &85.43 &\textbf{$+$3.33}  &87.44 &\textbf{$+$5.14} &84.85 &\textbf{$+$4.95} \\ RUM-II (cum) & & &   \\ \hline
CSM-III &\checkmark  &87.95 &\textbf{$+$3.62} &85.86 &\textbf{$+$3.76}  &87.26 &\textbf{$+$4.96} &85.34 &\textbf{$+$5.44} \\ RUM-I (max) & & &   \\ \hline
CSM-III &\checkmark  &87.44 &\textbf{$+$3.11} &85.33 &\textbf{$+$3.23}  &87.09 &\textbf{$+$4.79} &84.52 &\textbf{$+$4.62} \\ RUM-II (cum) & & &   \\ \hline
CSM-IV &\checkmark  &87.78 &\textbf{$+$3.45} &85.87 &\textbf{$+$3.77}  &88.12 &\textbf{$+$5.82} &85.60 &\textbf{$+$5.70} \\ RUM-I (max) & & &   \\ \hline
CSM-V &\checkmark  &87.09 &\textbf{+2.76} &84.93 &\textbf{$+$2.83}  &87.78 &\textbf{$+$5.48} &85.31 &\textbf{$+$5.41} \\ RUM-II (cum) & & &   \\ 

\hline 
\end{tabular}
\label{tab:tablecsmrum}
\end{table*}

\begin{figure}[h]
     \centering
     \begin{subfigure}[b]{0.85\textwidth}
         \centering
         \includegraphics[width=\textwidth]{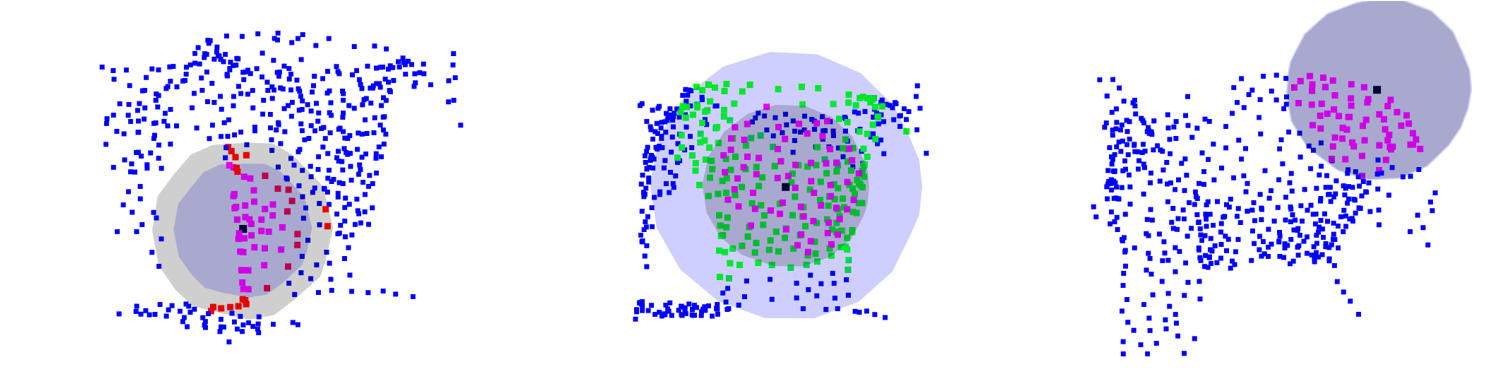}
         \caption{}
         \label{fig:rum_bg_a}
     \end{subfigure}
     \begin{subfigure}[b]{0.85\textwidth}
         \centering
         \includegraphics[width=\textwidth]{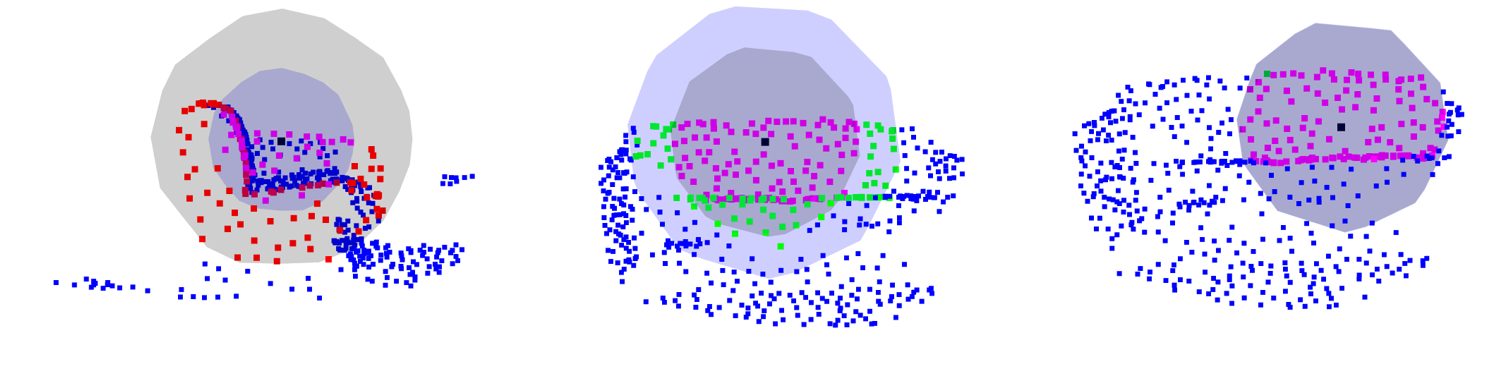}
         \caption{}
         \label{fig:rum_bg_b}
     \end{subfigure}
     \begin{subfigure}[b]{0.85\textwidth}
         \centering
         \includegraphics[width=\textwidth]{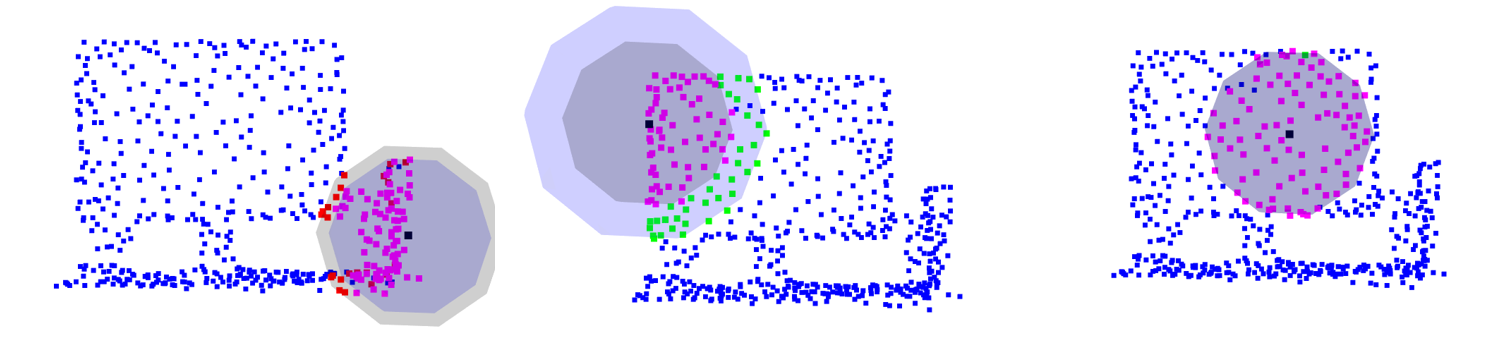}
         \caption{}
         \label{fig:rum_bg_c}
     \end{subfigure}
     \begin{subfigure}[b]{0.85\textwidth}
         \centering
         \includegraphics[width=\textwidth]{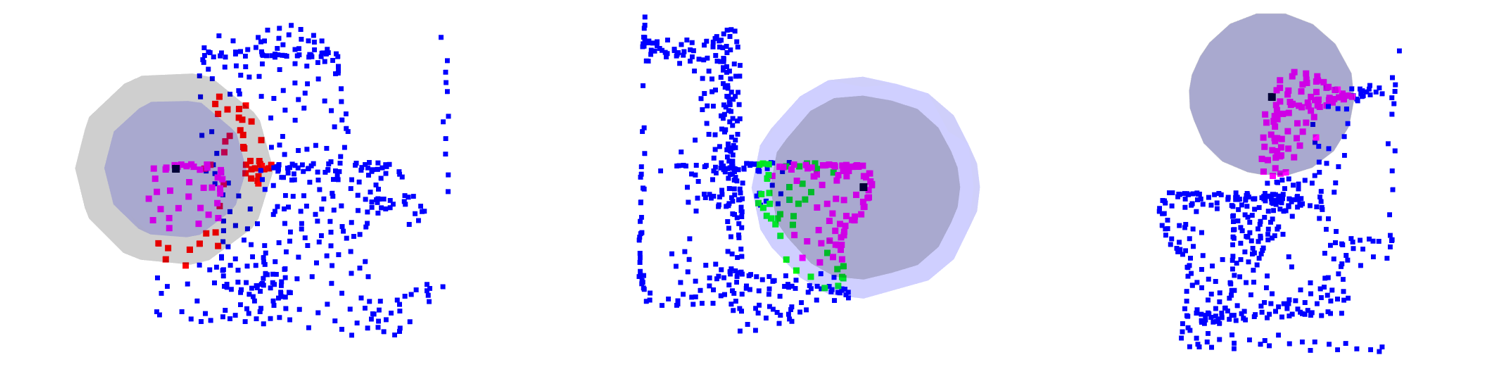}
         \caption{}
         \label{fig:rum_bg_d}
     \end{subfigure}

     \caption{The visual result of updated radius through RUM-II-(cum) applied only to the second layer. The objects belong to bin (a), display (b), sofa (c), and toilet (d) categories.}
     \label{fig:visualrumbg}
\end{figure}

Examples of local regions affected by RUM are given in Figure \ref{fig:rumvisual} for sample objects from OBJ\_ONLY data set. The first column shows sample regions where the radii are reduced by RUM-I-(max). The points excluded from the updated regions are marked with red color. We can observe that the regions shrink to fit a semantic part of the object while excluding the points not belonging to the part. For example, the radius of the region enclosing the leg of the chair is reduced to exlude points from the seat and base (Figure \ref{fig:rumvisual}a). Similarly, a region on the top of the desk is shrank to be confined to the planar region (Figure \ref{fig:rumvisual}b). In the second column, the opposite effect is demonstrated, where the radii are increased by RUM to include more points (marked with green) from the same semantic part of an object. Examples are the back of the chair and the side of the desk. The last column shows instances where the update is zero; hence the particular region did not change.

Among all RUM variations applied to the second layer only, RUM-I (max) yielded the largest increase in $MAcc$ as +4.03\% for the OBJ\_ONLY data. For the OBJ\_BG data, RUM-II (cum) resulted in the largest increase as +6.41\%. As in the case with CSM, RUM boosts the classification accuracy for OBJ\_BG data more than the OBJ\_ONLY data. 

In Figure \ref{fig:visualrumbg}, the effect of the RUM is demonstrated for sample objects from OBJ\_BG data set. We can observe a number of tendencies of the RUM while processing the noisy data, additional to the category-sensitive region adapting. One is to isolate salient object parts both from the object itself and also from irrelevant background. Examples are the excluded background points from the bin object in Figure \ref{fig:visualrumbg}a and the region shrinking to the back and seat of the sofa in Figure \ref{fig:visualrumbg}b. Another tendency of RUM is to isolate background regions from the object points as can be seen at the updated regions of display and toilet objects in Figure \ref{fig:visualrumbg}c and \ref{fig:visualrumbg}d. In the cases where RUM increased the region size for a center point (the second column of Figure \ref{fig:visualrumbg}) more points making the region distinct and salient for an object category are included in the region.

Guided by the results obtained by integrating CSM and RUM variations separately to PointNet++ architecture, we opted to apply them together to only to the second layer. We report in Table \ref{tab:tablecsmrum} the results of the experiments conducted by combining variations of CSM and RUM. The combination providing the highest increase in performance is CSM-I and RUM-I (max) for both OBJ\_ONLY and OBJ\_BG data. $MAcc$ increased by 5.25\% for OBJ\_ONLY data, and 6.75\% for OBJ\_BG, surpassing the increase obtained by integrating CSM or RUM separately. 

\begin{table*}[h]
\caption{Results of PointCNN with CSM on ScanObjectNN data set.}
\centering
\begin{tabular}{lccccccccc} \hline 
 & &\multicolumn{4}{c}{OBJ\_ONLY} &\multicolumn{4}{c}{OBJ\_BG}\\ \cmidrule(lr){3-6} \cmidrule(lr){7-10}
\textbf{Method} &3\textsuperscript{rd} &$Acc$ &+/- &$MAcc$ &+/- &$Acc$ &+/- &$MAcc$ &+/- \\ \hline

PointCNN & &85.5 &- &83.3 &- &86.1 &- &83.3 &-\\ \hline 
\multirow{1}{6em} {CSM-I} &\checkmark &87.03 &\textbf{$+$1.53} &85.55 &\textbf{$+$2.25} &88.75 &\textbf{$+$2.65} &86.10 &\textbf{$+$2.80} \\ 
\multirow{1}{6em} {CSM-II (sub)} &\checkmark &87.19 &\textbf{$+$1.69} &85.64 &\textbf{$+$2.34} &90.16 &\textbf{$+$4.06} &87.49 &\textbf{$+$4.19} \\
\multirow{1}{8em} {CSM-II (sum)}  &\checkmark &86.09 &\textbf{$+$0.59} &84.92 &\textbf{$+$1.62} &88.59 &\textbf{$+$2.49} &86.59 &\textbf{$+$3.29} \\
\multirow{1}{6em} {CSM-II (cat)}  &\checkmark &87.03 &\textbf{$+$1.53} &85.44 &\textbf{$+$2.14} &89.53 &\textbf{$+$3.43} &87.06 &\textbf{$+$3.76} \\ 
\multirow{1}{6em} {CSM-II (dot)}  &\checkmark &85.94 &\textbf{$+$0.44} &83.43 &\textbf{$+$0.13} &87.50 &\textbf{$+$1.40} &84.79 &\textbf{$+$1.49} \\ 
\multirow{1}{9em} {CSM-II (hadamard)}  &\checkmark &86.25 &\textbf{$+$0.75} &83.53 &\textbf{$+$0.23} &89.38 &\textbf{$+$3.28} &86.97 &\textbf{$+$3.67} \\  
\multirow{1}{6em} {CSM-III}  &\checkmark &87.19 &\textbf{$+$1.69} &86.11 &\textbf{$+$2.81} &88.91 &\textbf{$+$2.81} &86.62 &\textbf{$+$3.32} \\ 
\multirow{1}{6em} {CSM-IV}  &\checkmark &85.31 &$-$0.19 &83.98 &\textbf{$+$0.68} &86.56 &\textbf{$+$0.46} &84.01 &\textbf{$+$0.71} \\  

\hline 
\end{tabular}
\label{tab:tablecsm_pointcnn}
\end{table*}

\begin{table*}[h]
\caption{Results of PointCNN with RUM on ScanObjectNN data set.}
\centering
\begin{tabular}{lcccccccccc} \hline 
 & &\multicolumn{4}{c}{OBJ\_ONLY} &\multicolumn{4}{c}{OBJ\_BG}\\ \cmidrule(lr){3-6} \cmidrule(lr){7-10}
\textbf{Method} &3\textsuperscript{rd} &$Acc$ &+/- &$MAcc$  &+/- &$Acc$  &+/- &$MAcc$  &+/-\\ \hline
PointCNN  & &85.5 &- &83.3 &- &86.1 &- &83.3 &- \\ \hline 
\multirow{1}{8em} {RUM-I (cum)}  &\checkmark &87.83 &\textbf{$+$2.33} &85.80 &\textbf{$+$2.50} &89.33 &\textbf{$+$3.23} &86.20 &\textbf{$+$2.90} \\
\multirow{1}{8em} {RUM-I (max)}  &\checkmark &86.33 &\textbf{$+$0.83} &84.24 &\textbf{$+$0.94} &89.67 &\textbf{$+$3.57} &87.09 &\textbf{$+$3.79} \\
\multirow{1}{8em} {RUM-II (cum)}  &\checkmark &87.00 &\textbf{$+$1.50} &85.04 &\textbf{$+$1.74} &89.33 &\textbf{$+$3.23} &86.02 &\textbf{$+$2.72} \\ 
\multirow{1}{8em} {RUM-II (max)} &\checkmark &86.83 &\textbf{$+$1.33} &84.95 &\textbf{$+$1.65} &89.00 &\textbf{$+$2.90} &86.34 &\textbf{$+$3.04} \\  

\hline 
\end{tabular}
\label{tab:tablerum_pointcnn}
\end{table*}

\begin{table*}[!h]
\caption{Results of PointCNN  with merged combinations of CSM and RUM modules on ScanObjectNN data set.}
\centering
\begin{tabular}{lccccccccc} \hline 
 & &\multicolumn{4}{c}{OBJ\_ONLY} &\multicolumn{4}{c}{OBJ\_BG}\\ \cmidrule(lr){3-6} \cmidrule(lr){7-10}
\textbf{Method} &3\textsuperscript{rd} &$Acc$ &+/- &$MAcc$ &+/- &$Acc$ &+/- &$MAcc$ &+/- \\ \hline
PointCNN & &85.5 &- &83.3 &- &86.1 &- &83.3 &- \\ \hline 
CSM-II (sub)  &\checkmark &89.67 &\textbf{$+$4.17} &88.71 &\textbf{$+$5.41}  &90.50 &\textbf{$+$4.40} &87.89 &\textbf{$+$4.59} \\ RUM-I (cum) & & &  \\ \hline 
CSM-II (sub)  &\checkmark &88.67 &\textbf{$+$3.17} &87.99 &\textbf{$+$4.69}  &90.83 &\textbf{$+$4.73} &89.08 &\textbf{$+$5.78} \\ RUM-II (cum) & & &  \\ \hline 
CSM-III  &\checkmark &87.83 &\textbf{$+$2.33} &86.84 &\textbf{$+$3.54}  &88.83 &\textbf{$+$2.73} &86.24 &\textbf{$+$2.94} \\ RUM-I (cum) & & &  \\ \hline 
CSM-III  &\checkmark &87.33 &\textbf{$+$1.83} &86.01 &\textbf{$+$2.71}  &88.83 &\textbf{$+$2.73} &86.21 &\textbf{$+$2.91} \\ RUM-II (cum) & & &  \\ 
\hline 
\end{tabular}
\label{tab:tablecsmrum_pointcnn}
\end{table*}

\subsubsection{Results with CSM and RUM integrated to PointCNN}

We observed that CSM and RUM provided superior performance boost when integrated only to the second layer of PointNet++. The second layer is the one where largest local regions are formed before extraction of a global feature for the entire point cloud. Using this observation, we integrated variations of CSM and RUM only to the last $\mathcal{X}$-Conv layer of PointCNN. 

With PointCNN, we excluded CSM-V from the analysis, since its time complexity is quite high (see Table \ref{tab:complexity}) and its performance boost does not compensate for this disadvantage as compared to the other variations (see Table \ref{tab:tablecsm}). The classification results obtained with integration of variations of CSM to the third $\mathcal{X}$-Conv of PointCNN are given in Table \ref{tab:tablecsm_pointcnn}. The first row includes results yielded by PointCNN without integration of CSM. With the exception of CSM-IV, all variations of CSM increased the performance on the OBJ\_ONLY data. The highest performance increase on the OBJ\_ONLY set in terms of $MAcc$ is $2.81$\% with CSM-III. The contribution of CSM-II (sub) is close with $2.34$\% increase. Both variations, CSM-III and CSM-II (sub), achieved the highest boost in terms of $Acc$ with $1.69$\% increase.

On the OBJ\_BG set, all CSM variations integrated to PointCNN achieved even higher performance increase, consistent with the results obtained with PointNet++. These results are another indicator of the suitability of CSM for classifying objects with cluttered background. Among the variations, CSM-II (sub) provided the highest performance increase in terms of both $Acc$ ($+4.06$\%) and $MAcc$ ($+4.19$\%).

The classification results obtained with integration of variations of RUM to the third $\mathcal{X}$-Conv layer of PointCNN are given in Table \ref{tab:tablerum_pointcnn}. All variations of RUM yielded an increase in classification performance. On the OBJ\_ONLY set, the best performing variations are RUM-I (cum) and RUM-II (cum), with $2.50$\% and $1.74$\% increase in $MAcc$, respectively. Similar to CSM variations, RUM variations  achieved higher performance boost on the OBJ\_BG set. All variations yielded an increase on $MAcc$ higher than $2.7$\%, with RUM-I (max) providing the highest increase as $+3.79$\%. However, the performance increase of RUM-I (max) is much lower on the OBJ\_ONLY set as compared to the other variations. RUM-I (cum) and RUM-II (cum) are the variations that consistently provided high performance increase on both OBJ\_ONLY and OBJ\_BG sets.

Observing that CSM-II (sub), CSM-III, RUM-I (cum) and RUM-II (cum) achieved high performance increase on both OBJ\_ONLY and OBJ\_BG sets, we decided to integrate merged combinations of them to the third $\mathcal{X}$-Conv layer of PointCNN. The results are given in Table \ref{tab:tablecsmrum_pointcnn}. Although merging of CSM-III with RUM-I (cum) or RUM-II (cum) increased the performance with respect to the default values obtained with PointCNN, it did not result in performance increase on OBJ\_BG data as compared to applying CSM-III alone. However, the performance boost increased for both data sets when we merged CSM-II (sub) with RUM-I (cum) or RUM-II (cum) as compared to their isolated applications. For example, with merging of CSM-II (sub) and RUM-II (cum), the increase in $MAcc$ is $5.78$\%, while the increase is $4.19$\% with CSM-II (sub) only, and $2.72$\% with RUM-II (cum) only.

To summarize, the highest classification performance on the OBJ\_ONLY set is obtained with PointCNN integrated with CSM-II (sub) and RUM-I (cum), yielding $89.67$\% $Acc$ and $88.71$\% $MAcc$. For the OBJ\_BG set, PointCNN integrated with CSM-II (sub) and RUM-II (cum) achieved the highest performance with $90.50$\% $Acc$ and $87.89$\% $MAcc$.

\subsection{Results on ShapeNet}

We conducted additional experiments to evaluate the effectiveness of the center shift and radius update modules on another 3D object data set, the ShapeNet. We selected CSM-II (sub) and RUM-II (cum), as we observed that they consistently yielded good results. The classification performance of PointCNN and PointNet++ with the integration of the two modules are provided in Table \ref{tab:tablecsmrum_shapenet}. The first row gives $Acc$ and $MAcc$ values of the default versions of PointNet++ and PointCNN architectures. The second and third rows contain the results obtained with individual integration of CSM-II (sub) and RUM-II (cum), respectively. In the last row, the results achieved with the merged integration of the two modules are presented.

 \begin{table*}[!h]
\caption{Results of PointCNN and PointNet++ with the integration of CSM and RUM on ShapeNet data set.}
\centering
\begin{tabular}{lcccccccc} \hline 
  &\multicolumn{4}{c}{PointNet++} &\multicolumn{4}{c}{PointCNN}\\ \cmidrule(lr){2-5} \cmidrule(lr){6-9}
\textbf{Method}  &$Acc$ &+/- &$MAcc$ &+/- &$Acc$ &+/- &$MAcc$ &+/- \\ \hline
Baseline  &98.47 &- &95.94 &- &98.58 &- &94.92 &- \\ \hline 
CSM-II (sub)  &98.78 &\textbf{$+$0.31} &97.40 &\textbf{$+$1.46}  &98.85 &\textbf{$+$0.27} &96.92 &\textbf{$+$2.00} \\ \hline
RUM-II (cum)  &98.64 &\textbf{$+$0.17} &96.88 &\textbf{$+$0.94}  &98.99 &\textbf{$+$0.41} &96.09 &\textbf{$+$1.17} \\ \hline 
CSM-II (sub) &98.54 &\textbf{$+$0.07} &96.90 &\textbf{$+$0.96}  &99.10 &\textbf{$+$0.52} &97.31 &\textbf{$+$2.39} \\  \& RUM-II (cum) & & &  \\ \hline 

\end{tabular}
\label{tab:tablecsmrum_shapenet}
\end{table*}

With PointNet++, the isolated applications of CSM-II (sub) and RUM-II (cum) introduced $1.46$\% and $0.94$\% increase in $MAcc$, respectively. However, their merged application did not improve the results as compared to the case with the integration of only CSM-II (sub).

With PointCNN, the modules are more effective. CSM-II (sub) and RUM-II (cum) increased the $MAcc$ values by $2.00$\% and $1.17$\%, respectively. When the two modules are jointly applied, a $2.39$\% increase in $MAcc$ is obtained.

While ScanObjectNN is composed of 3D point clouds captured from real objects, ShapeNet is a CAD model dataset. The results indicate that our proposed modules are effective on 3D scans of real objects as well as on 3D models created with CAD.

\subsection{Time and Space Complexity Analysis}
The number of parameters as a measure of space complexity and the number of floating-point operations per sample (FLOPs) as a measure of time complexity are presented in Table \ref{tab:complexity}. The initial row of the table outlines the time and space complexities of the baseline architectures, PointNet++ and PointCNN. In each of the following rows, number of parameters and FLOPs of the overall architecture when integrated with the corresponding module are given.

Both CSM and RUM increase the time and space complexity of the architectures substantially, especially of PointCNN. Our comments addressing this disadvantage is given in Section \ref{section:discussion}.  In terms of the number of parameters, there is not much difference between variations of either CSM or RUM. Variations of RUM are similar in terms of FLOPs. Among the variations of CSM, CSM-V requires far more operations. Also, CSM-V does not bring a significant performance boost compared to the other versions (see Table \ref{tab:tablecsm}).

\begin{table*}[!h]
\caption{Time and space complexity of CSM and RUM modules. The modules are integrated to the second layer of PointNet++ and the third layer of PointCNN frameworks.}
\centering
\begin{tabular}{lcccc} \hline 
&\multicolumn{2}{c}{PointNet++} &\multicolumn{2}{c}{PointCNN}\\ \cmidrule(lr){2-3} \cmidrule(lr){4-5}
 \textbf{Method}  &\#params &FLOPs &\#params &FLOPs  \\ \hline
 Baseline  &1.46M &1686M &0.28M &197M  \\\hline
 CSM-I   &1.55M &2331M &3.44M &418M  \\ 
 CSM-II (sub)  &1.57M &2586M &3.54M &573M  \\ 
 CSM-II (sum)  &1.57M &2586M &3.54M &573M  \\ 
 CSM-II (cat)  &1.58M &2719M &3.59M &648M  \\ 
 CSM-II (dot)  &1.56M &2453M &3.49M &499M  \\ 
 CSM-II (hadamard)  &1.57M &2586M &3.54M &573M  \\ 
 CSM-III  &1.58M &2349M &3.63M &440M  \\ 
 CSM-IV  &1.59M &5078M &3.67M &593M  \\ 
 CSM-V  &1.47M &21968M &2.89M &7331M  \\
 RUM-I-cum  &1.48M &3929M &1.61M &714M  \\
 RUM-I-max  &1.48M &3920M &1.61M &711M  \\
 RUM-II-cum  &1.53M &4030M &1.64M &743M  \\
 RUM-II-max  &1.53M &4021M &1.64M &740M  \\
 
\hline 
\end{tabular}
\label{tab:complexity}
\end{table*}

\section{Discussion}
\label{section:discussion}

Most of the variations of the modules result in an increase in classification performance. However, CSM-V brings a high cost to the architectures in terms of time complexity. Therefore, we exclude CSM-V from our discussion on variations of the modules.

The level of contribution of CSM variations changes depending on the data set and on the architecture to which they are integrated. With both architectures, CSM-III tends to give higher performance boost on the OBJ\_ONLY data, where no background clutter is present. They are closely followed by CSM-I and versions of CSM-II. With PointCNN, CSM-III and CSM-II (sub) give comparable results on the OBJ\_ONLY set.

When tested on the OBJ\_BG data, versions of CSM-II take over. With PointNet++, CSM-II (sum) performs best on the OBJ\_BG set, while with PointCNN, CSM-II (sub) achieves the highest classification accuracy. CSM-II (sub) consistently improves the performance for both architectures and data sets.

Relying on these results, we recommend CSM-II (sub) as the first choice among the CSM variations. It updates each center point by taking into account both feature similarities and positional relations when calculating the attention weights within the examined region. It tends to perform well for object instances with or without background clutter. CSM-III is a good alternative for uncluttered data. Since center points from close regions contribute to the calculation of the shift vector in CSM-III, it might be adversly affected by the noise in the background.

All of the four RUM variations when applied to the second layer of PointNet++ and to the third layer of PointCNN improved the classification accuracy. With PointNet++, RUM-II (cum) provided the highest increase on the performance in terms of $MAcc$ for OBJ\_ONLY and OBJ\_BG sets. With PointCNN, the top performing variation for OBJ\_BG set is RUM-I (max), however its contribution is much lower on OBJ\_ONLY set compared to other versions. RUM-I (cum) and RUM-II (cum) consistently achieved high performance increase on both sets of ScanObjectNN. As a result of these observations, we recommend RUM-II as the first choice of the RUM variations.

As stated before, the variations of the modules behave differently depending on the data set and the main architecture. The interaction between the CSM and RUM modules when they are merged is complex. Some combinations perform worse than individual components. We would recommend to first merge the versions that individually perform best among the alternatives. In many of the cases, their joint application improves the classification performance.

An important issue is the added computational complexity with the employment of the proposed modules. Although the boost in $MAcc$ they bring for OBJ\_BG data of ScanObjectNN reaches $+6.75$ with PointNet++, and $+5.78$ with PointCNN, alleviating the computational burden would be preferable. The modularity of these structures is an advantage; however, it contributes to the heightened number of parameters and operations. Region descriptors are extracted twice, once for inferring the locations and sizes of the receptive fields, and once for feature extraction and aggregation from these local regions. As future work, we investigate designing attention-based architectures where local region-learning and feature aggregation within the region are jointly performed.

\section{Conclusion}
\label{section:conclusion}
In this work, we proposed variations of two modules, Central Shift Module and Radius Update Module to learn to infer the positions and radii of local regions organized in hierarchical 3D point-based deep learning architectures. We integrated variations of CSM and RUM to the classification networks of PointNet++ and PointCNN, and demonstrated their effectiveness on the publicly available ScanObjectNN and ShapeNet data sets. We observed that applying all variations of these modules to the second layer of PointNet++, where semantic regions are formed, increased the classification performance significantly. Integrating CSM and RUM separately only to the second layer of PointNet++ resulted in 6.17\% and 6.41\% increase in $MAcc$, respectively, for OBJ\_BG data of the ScanObjectNN data set, while combining both modules yielded a performance increase of 6.75\%. 
The integration of both modules to the third $\mathcal{X}$-Conv layer of PointCNN yielded 5.78\% in $MAcc$ on OBJ\_BG set of ScanObjectNN.  On the ShapeNet data set, the application of CSM and RUM modules increased the $MAcc$ of PointNet++ and PointCNN by 0.96 and 2.39, respectively. These results demonstrate that the proposed tools for learning the location and size of the local regions, hence the receptive fields, within the task network through minimization of the task loss are effective at improving data organization for 3D point-based deep networks.

\section*{Acknowledgements}
The authors acknowledge the support of The Scientific and Technological Research Council of Turkey (TUBITAK), Project No: 121E088.

\section*{Declaration of Competing Interest}
The authors declare that they have no known competing financial interests or personal relationships that could have appeared to influence the work reported in this paper.

\bibliographystyle{elsarticle-num-names}
\bibliography{ref.bib}
\end{document}